\documentclass[letterpaper]{article} 
\usepackage{aaai2026}  
\usepackage{times}  
\usepackage{helvet}  
\usepackage{courier}  
\usepackage[hyphens]{url}  
\usepackage{graphicx} 
\usepackage{natbib}  
\usepackage{caption} 
\usepackage{amsmath}
\usepackage{amssymb}
\usepackage{multirow}
\usepackage{makecell}
\usepackage{verbatim}
\usepackage{algorithm}
\usepackage{algorithmic}

\usepackage{newfloat}
\usepackage{listings}
\DeclareCaptionStyle{ruled}{labelfont=normalfont,labelsep=colon,strut=off} 
\floatstyle{ruled}
\newfloat{listing}{tb}{lst}{}
\floatname{listing}{Listing}
\title{Temporal Inconsistency
Guidance for Super-resolution Video
Quality Assessment}
\author{
    Yixiao Li\textsuperscript{\rm 1},
    Xiaoyuan Yang\textsuperscript{\rm 1}\equalcontrib, 
    Weide Liu\textsuperscript{\rm 2}, 
    Xin Jin\textsuperscript{\rm 3}, 
    Xu Jia\textsuperscript{\rm 4}, 
    Yukun Lai\textsuperscript{\rm 5},  
    Paul L. Rosin\textsuperscript{\rm 5}, 
    Hantao Liu\textsuperscript{\rm 5},
    Wei Zhou\textsuperscript{\rm 5}\equalcontrib
}
\affiliations{
    \textsuperscript{\rm 1}The School of Mathematical Sciences, Beihang University, Beijing 100191, China\\
    \textsuperscript{\rm 2}Harvard Medical School, Harvard University, USA\\
    \textsuperscript{\rm 3}The College of Info Science $\&$ Technology, Eastern Institute of Technology (EIT), Ningbo, China\\
    \textsuperscript{\rm 4}The School of Artificial Intelligence, Dalian University of Technology, Dalian 116081, China
    \textsuperscript{\rm 5}the School of Computer Science and Informatics, Cardiff University, Cardiff CF24 4AG, United Kingdom\\
    18335310648@163.com, xiaoyuanyang@vip.163.com, weide001@e.ntu.edu.sg, jinxin@eitech.edu.cn, xjia@dlut.edu.cn, laiy4@cardiff.ac.uk, rosinpl@cardiff.ac.uk, liuh35@cardiff.ac.uk. zhouw26@cardiff.ac.uk
}

\begin{document}

\maketitle

\begin{abstract}
As super-resolution (SR) techniques introduce unique distortions that fundamentally differ from those caused by traditional degradation processes (e.g., compression), there is an increasing demand for specialized video quality assessment (VQA) methods tailored to SR-generated content. One critical factor affecting perceived quality is temporal inconsistency, which refers to irregularities between consecutive frames. However, existing VQA approaches rarely quantify this phenomenon or explicitly investigate its relationship with human perception. Moreover, SR videos exhibit amplified inconsistency levels as a result of enhancement processes. In this paper, we propose \textit{Temporal Inconsistency Guidance for Super-resolution Video Quality Assessment (TIG-SVQA)} that underscores the critical role of temporal inconsistency in guiding the quality assessment of SR videos. We first design a perception-oriented approach to quantify frame-wise temporal inconsistency. Based on this, we introduce the Inconsistency Highlighted Spatial Module, which localizes inconsistent regions at both coarse and fine scales. Inspired by the human visual system, we further develop an Inconsistency Guided Temporal Module that performs progressive temporal feature aggregation: (1) a consistency-aware fusion stage in which a visual memory capacity block adaptively determines the information load of each temporal segment based on inconsistency levels, and (2) an informative filtering stage for emphasizing quality-related features. Extensive experiments on both single-frame and multi-frame SR video scenarios demonstrate that our method significantly outperforms state-of-the-art VQA approaches. The code is publicly available at \url{https://github.com/Lighting-YXLI/TIG-SVQA-main}.
\end{abstract}


\begin{figure}[t]
\centerline{\includegraphics[width=0.95\linewidth]{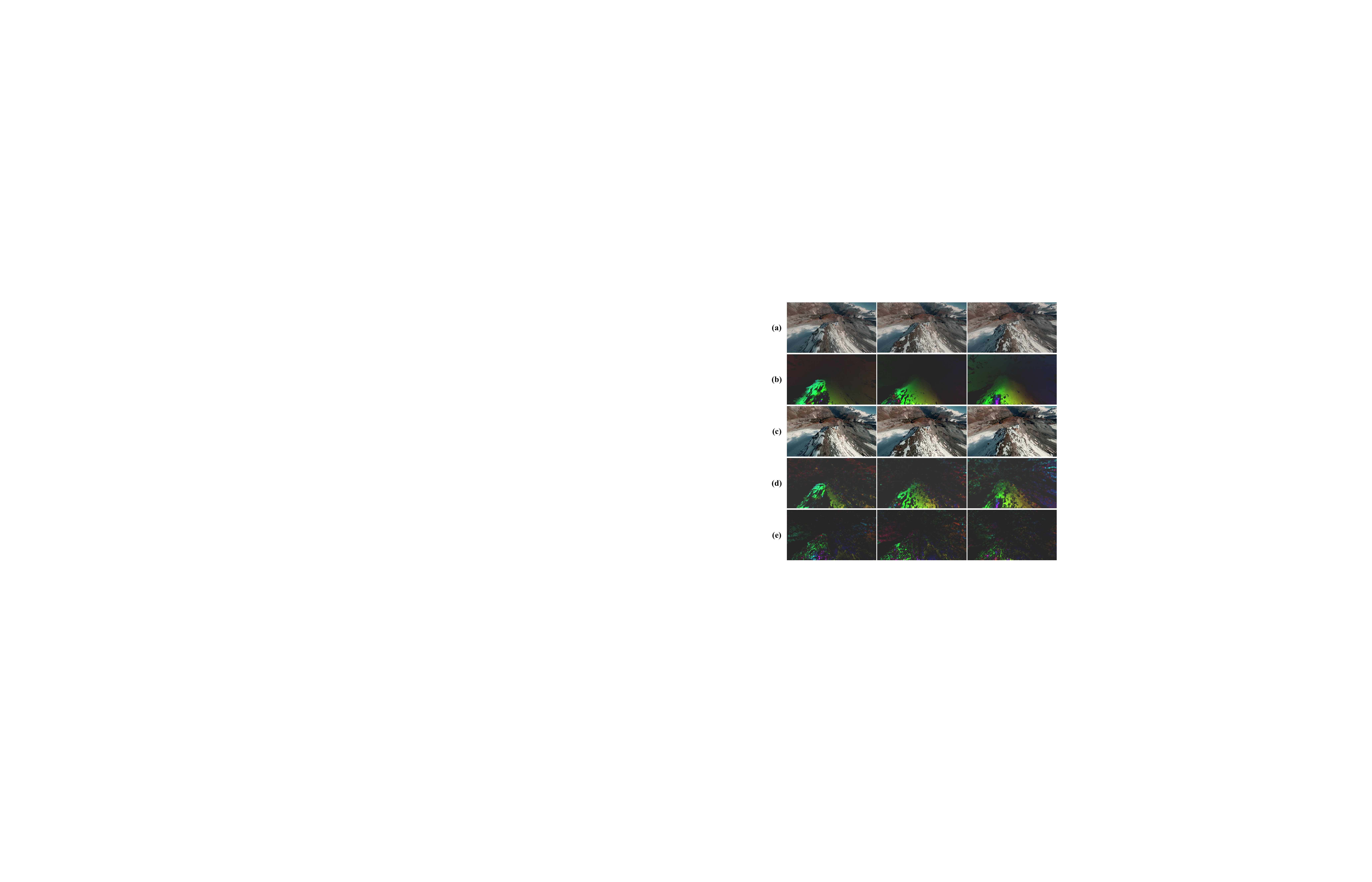}}
\caption{Visualizations for comparing temporal inconsistency with motion. Rows (a) and (c) show consecutive frames of a reference video and SR video, respectively. (b) and (d) are the optical flow of (a) and (c), respectively. (e) is the \textbf{temporal inconsistency} information for (c).}
\label{fig1}
\end{figure}

\section{Introduction}
\label{sec:intro}
The rapid advancement of video processing and transmission technologies has led to an explosion of diverse video content, profoundly influencing daily life. Numerous VQA datasets have been developed, each focusing on different types of distortions such as compression distortions, user-generated degradations, and super-resolution (SR) artifacts~\cite{konvid-1k, LIVEVQC, xu2021perceptual,VSRQAD}. Among them, SR introduces a unique class of artifacts, including hallucinated textures and temporal flickering, which differ substantially from traditional distortions. Although several SR-oriented VQA datasets have recently emerged, the development of SR-specific VQA models remains a pressing challenge due to the distinctive characteristics of SR-generated content.

\begin{figure}[t]
    \begin{minipage}[b]{0.45\linewidth}
        \includegraphics[width=\linewidth]{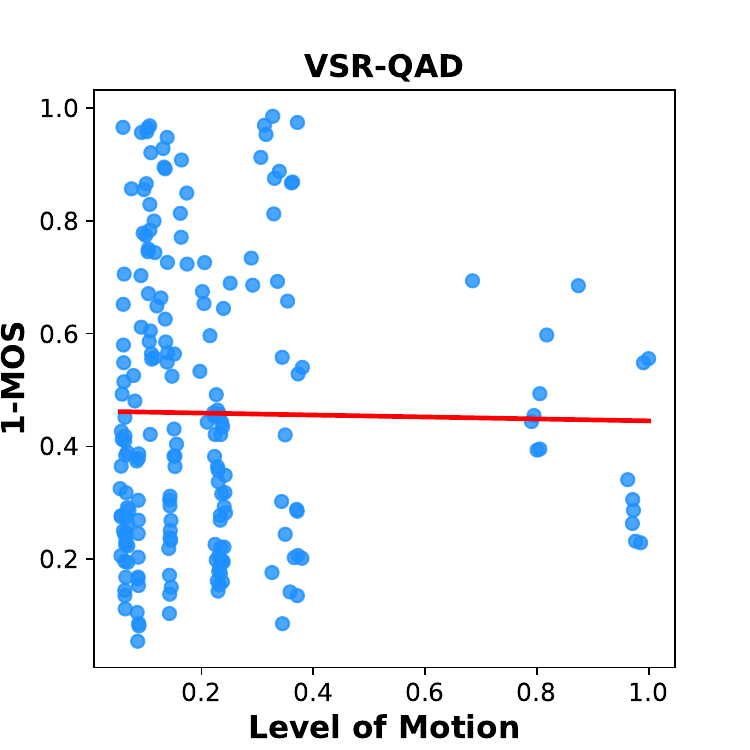} 
    \end{minipage}
    \hfill
    \begin{minipage}[b]{0.45\linewidth}
        \includegraphics[width=\linewidth]{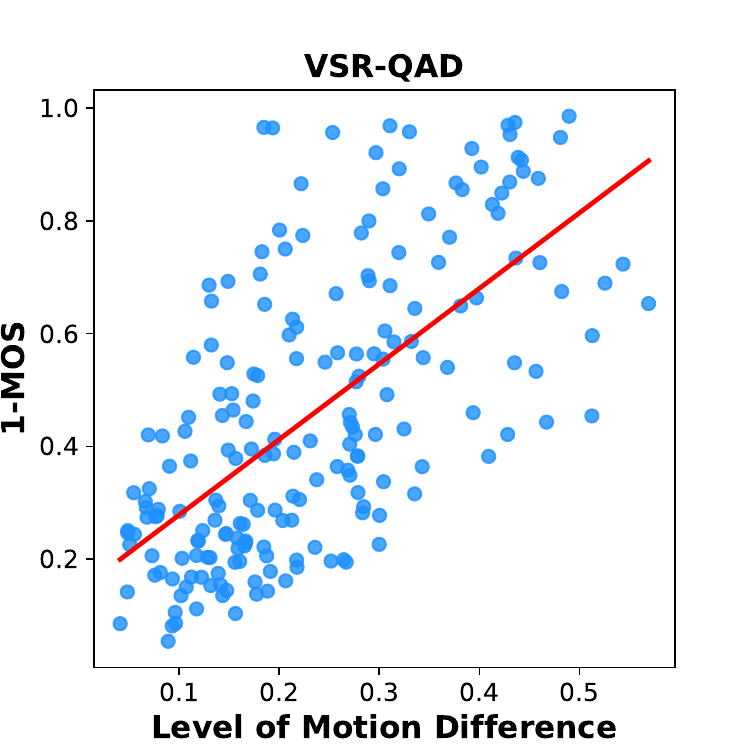} 
    \end{minipage}

    \caption{Correlation and performance comparison between motion and motion difference for model guidance. We analyze the correlation (SRCC) between perceptual quality (1–MOS) and motion or motion difference based on video complexity. Performance comparison shows that motion achieves SRCC/PLCC of 0.885/0.913, while motion difference reaches 0.939/0.942 on the Combined-VSR dataset.}
    \label{fig2}
\end{figure}

Due to the increasing demand for evaluating the quality of various video distortions, a wide range of VQA methods~\cite{CHIPQA,GSTVQA,LAGTVQA,SIMPLEQA,STFR,STIVQA,TLVQA,VIDEVAL,VMVQA} have gained significant attention. Among the critical factors influencing video quality, temporal inconsistency refers to irregularities or disruptions in dynamic scenes (such as motion artifacts, abrupt transitions, or unnatural visual changes) that deviate from the expected smooth flow of visual content across consecutive frames. It plays a pivotal role. Recent VQA methods typically model temporal relationships using techniques such as frame differencing~\cite{STIVQA,SpEED-QA}, optical flow analysis~\cite{STIVQA}, temporal slicing~\cite{VSRQAD,CHIPQA}, natural scene statistics~\cite{BLIINDS,VSFA}, and 3D convolutional neural networks~\cite{WILDBVQA,MBVQA} applied directly to distorted videos. However, none of these approaches explicitly quantify temporal inconsistency levels or examine their correlation with human perception. Moreover, the rapid advancement of SR technologies~\cite{SR,VSR} has introduced additional challenges, as enhancement processes often amplify temporal inconsistencies, making it even more critical to address this issue in SR-specific VQA.

Given the potential to explore temporal inconsistency in SR-specific VQA, we begin by quantifying this phenomenon and investigating its correlation with perceptual quality. Temporal inconsistency in SR videos often arises during motion transitions~\cite{zhang2020there}, prompting us to analyze the relationship between motion complexity and the mean opinion score (MOS). As shown in the left part of Figure~\ref{fig2}, motion complexity exhibits a weak correlation with perceptual quality, likely due to its strong dependence on the intrinsic complexity of scene content. This suggests that scene content can mask temporal artifacts, making motion-based signals less reliable. To mitigate this masking effect, we compute the difference in motion information between the distorted video and its reference, which we refer to as the temporal inconsistency information. As illustrated in the right part of Figure~\ref{fig2}, the temporal inconsistency correlates strongly with perceptual quality, reinforcing the value of temporal inconsistency as a perceptual cue. Visualizations of this temporal inconsistency information are further provided in Figure~\ref{fig1}.

Therefore, incorporating temporal inconsistency information as a guiding signal in SR-specific VQA may be beneficial for improving alignment with human perceptual preferences. To this end, we propose the \textbf{Temporal Inconsistency Guidance for Super-resolution Video Quality Assessment (TIG-SVQA)} framework, which integrates temporal inconsistency information to guide both spatial and temporal feature modeling.

To process spatial information, we highlight frame-by-frame inconsistent regions using the computed temporal inconsistency map. Spatial features are extracted at two granularities. At the coarse level, we introduce a Deformable Window Super-Attention (DW-SA) Transformer to capture large-scale inconsistencies caused by major scene transitions or fast motion, leveraging the global modeling capability of Transformers~\cite{Swin}. At the fine level, we adopt CNNs to detect subtle distortions arising from minor motion or slow transitions, utilizing their strength in capturing local details~\cite{resnet}. The two levels of features are subsequently fused to form the final spatial representation.

For temporal modeling, we design a two-stage aggregation including Consistency-aware Fusion and Informative Filtering. The first stage is inspired by the visual working memory (VWM) mechanism in human perception. While previous works~\cite{VMVQA,VSFA} have considered VWM, they often ignore its capacity limitation, a key property supported by cognitive studies~\cite{MC1,MC2,MC3,MC4}. To address this, we propose a visual memory capacity block that dynamically allocates memory resources across time segments based on the inconsistency intensity. In the second stage, we perform temporal informative feature selection to retain quality-related features, ultimately enabling perceptually aligned, cross-time-scale quality prediction.

The main contributions of this paper are summarized as follows:

1. We propose the Temporal Inconsistency Guidance for Super-resolution Video Quality Assessment (TIG-SVQA) method, developing temporal inconsistency guidance for quality prediction and validating both its rationale and effectiveness.

2. We propose the Inconsistency Highlighted Spatial Module (IHSM), designed to decouple temporal inconsistency and highlight pixel-level regions exhibiting temporal irregularities across two spatial granularities.

3. We propose the Inconsistency Guided Temporal Module (IGTM), which includes the consistency-aware fusion stage and the informative filtering stage. In particular, a visual memory capacity block is proposed to dynamically allocate memory thresholds for temporal feature segmentation based on detected inconsistency levels.

\begin{figure*}[t]
\centerline{\includegraphics[width=\linewidth]{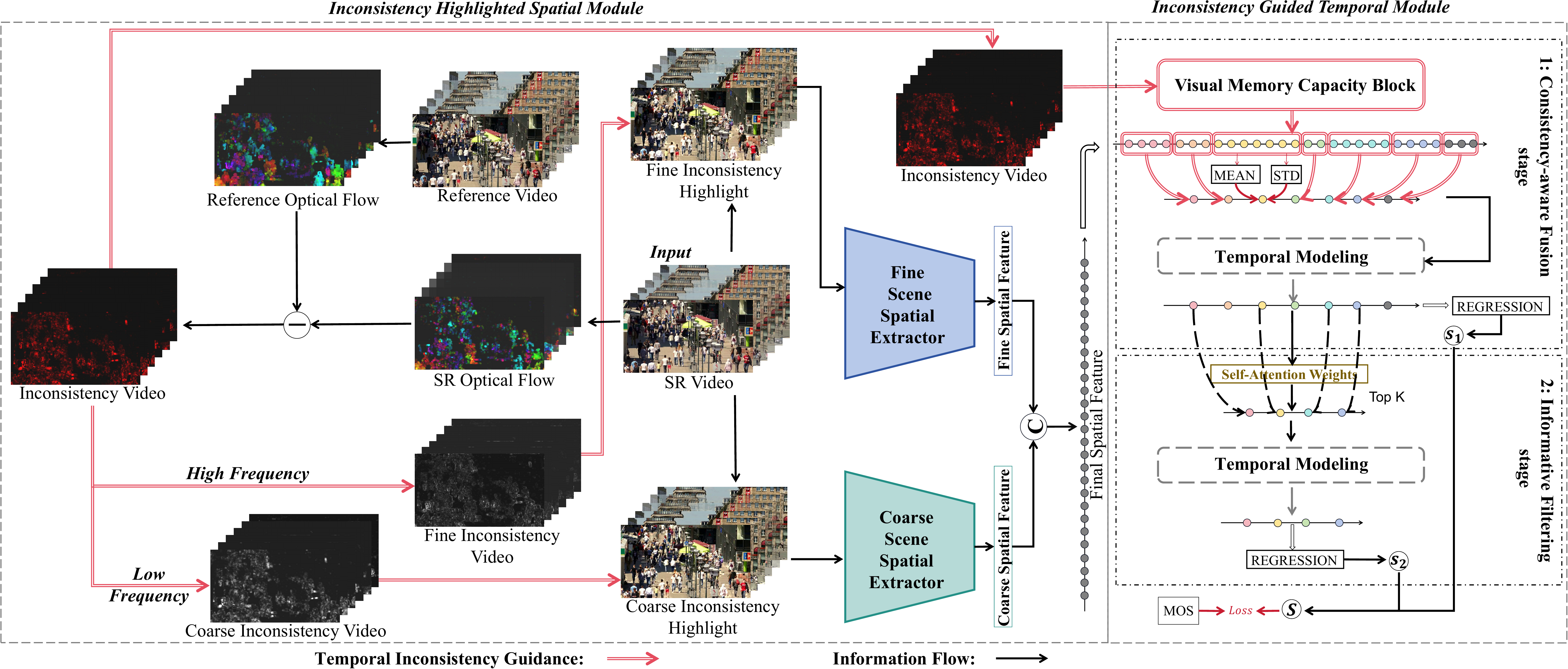}}
\caption{The framework of the proposed TIG-SVQA. The spatial module computes temporal inconsistency and applies pixel-level weighting at both coarse and fine granularities to emphasize inconsistent regions. The temporal module involves two stages: Consistency-aware Fusion and Informative Filtering.}
\label{fig3}
\end{figure*}

\section{Related Work}
\subsection{Temporal Relationship Modeling in VQA}
Contemporary VQA methods have actively explored feature extraction strategies for modeling temporal relationships. These approaches can be broadly categorized into five types: frame difference~\cite{STIVQA,SpEED-QA}, optical flow~\cite{STIVQA}, spatio-temporal slicing~\cite{VSRQAD,CHIPQA}, 3D-CNNs~\cite{WILDBVQA,MBVQA}, and multi-level methods~\cite{BLIINDS,TLVQA}. However, many VQA models rely heavily on handcrafted features to capture temporal features. Specifically, BLIINDS~\cite{BLIINDS} designs features to assess motion coherence and global motion (egomotion). TLVQM~\cite{TLVQA} employs low- and high-level frame complexity and uses the standard deviation of temporal features. Recently, there has been a shift towards learning-based approaches, such as STI-VQA~\cite{STIVQA}, MBVQA~\cite{MBVQA}, and FAST-VQA~\cite{fastvqa}. However, these methods are not designed for SR scenarios. As a result, their performance becomes unstable in SR scenarios, where temporal inconsistencies are often amplified by the upsampling process. 

As for SR-specific VQA, few studies such as ERQA~\cite{kirillova2021erqa}, Zhou et al.'s SR-VSR database and model~\cite{VSRQAD}, and the SR-VQA model~\cite{cao2024sr}, have proposed CNN- and saliency-based approaches tailored to SR video scenarios. However, the field still has substantial room for improvement, particularly in exploring temporal inconsistency guidance.
\subsection{Visual Working Memory Mechanism in VQA}
Recent research in psychology has highlighted the crucial role of visual working memory (VWM) in shaping temporal perception~\cite{VM1,VM2}. However, its application in VQA remains relatively underexplored. Neuroscientific studies have shown that memory performance improves with increased salience of visual stimuli~\cite{VM3}. Building on this, VM-VQA~\cite{VMVQA} employs saliency maps derived from Complete Local Binary Patterns on residual frames to model visual memory through statistical analysis. 

A defining characteristic of VWM is its limited storage capacity, which constrains both the retention and processing of visual information~\cite{VM1,MC2}. Typically, this capacity is restricted to approximately 3 to 7 visual objects~\cite{MC1,MC2}, and memory accuracy tends to decline as object complexity increases~\cite{MC3,MC4}. Further research has shown that VWM dynamically allocates its limited resources by prioritizing salient or critical scenes, which directly impacts how temporal relation is processed and modeled~\cite{MC5,MC6}. 

\begin{figure*}[t]
\centerline{\includegraphics[scale=0.42]{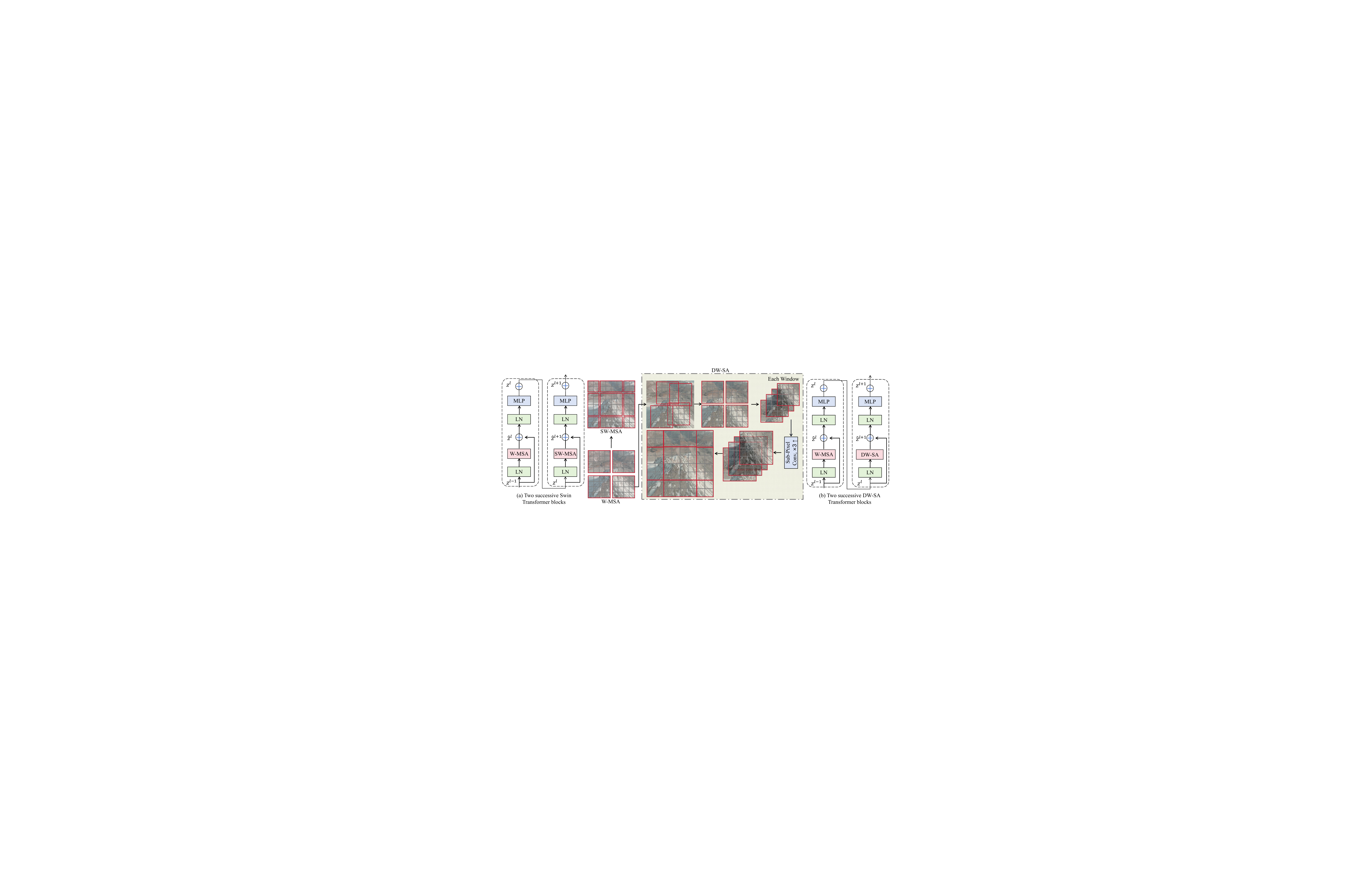}}
\caption{The details of the proposed Deformable Window Super-Attention (DW-SA) Transformer block, which adaptively adjusts window locations, up-samples features within each window, and then shifts the windows.}
\label{fig4}
\end{figure*}

\section{Proposed Method}
\label{sec3}
In this work, we propose a \textbf{Temporal Inconsistency Guidance for Super-resolution Video Quality Assessment} method. As illustrated in Figure \ref{fig3}, the temporal inconsistency information guides the model learning in both spatial and temporal processing.

\subsection{Inconsistency Highlighted Spatial Module (IHSM)}
\label{sec31}
In this section, we introduce the inconsistency guidance for the spatial dimension. We first calculate the temporally inconsistent areas of SR videos. Taking a pair of SR video and the corresponding reference $\rm V_{D}$, $\rm V_{R}$$\rm \in \mathbb{R}^{F\times W\times H\times 3}$ as inputs, where $\rm F$ is the total number of frames, and $\rm W, H, 3$ denote the width, height, and number of channels of each frame, respectively. The temporal inconsistency information $\rm V_{I}$ is captured as follows:
\begin{equation}
\small
\rm
V_{I}= \|(\rm OF(V_{R})-OF(V_{D}))\|_{2}, V_{D} \in \text{SR videos},
\label{eq1}
\end{equation}
where $\rm OF(\cdot)$ refers to optical flow computation. $\|(\cdot)\|_{2}$ is 2-norm. 

To capture temporal inconsistencies at different scales, we decouple the inconsistency information into coarse and fine granularities. The \textbf{coarse-grain} focuses on fast-changing regions such as scene cuts or rapid motions, obtained by applying a Gaussian low-pass filter to the video $\rm V_{I}$ in the frequency domain:
\begin{equation}
\small
\rm
V_{I}^{C} = \mathcal{F}^{-1}(H_{L} \cdot \mathcal{F}(V_{I})),
\end{equation}
where $\rm \mathcal{F}$ and $\rm \mathcal{F}^{-1}$ denote the Fourier and inverse Fourier transforms, and $\rm H_{L}$ is a Gaussian low-pass filter with a cutoff frequency set to $5\%$ of the frame’s longer dimension~\cite{hlpass}.

The \textbf{fine-grain} captures subtle inconsistencies from slow transitions or minor motions by applying a Gaussian high-pass filter:
\begin{equation}
\small
\rm
V_{I}^{F} = \mathcal{F}^{-1}(H_{H} \cdot \mathcal{F}(V_{I})), \quad H_{H} = 1 - H_{L}.
\end{equation}

Both filtered outputs are normalized and used to highlight inconsistency regions in the SR video $V_{D}$:
\begin{equation}
\small
\rm \{\hat{V_{D}}^{C},\hat{V_{D}}^{F}\} = \text{Norm}(\{V_{I}^{C},V_{I}^{F}\}) \times V_{D} + V_{D},
\end{equation}
where $\text{Norm}(\cdot)$ scales values to $[0, 1]$, and the outputs $\rm \hat{V_{D}}^{C}$ and $\rm \hat{V_{D}}^{F}$ are normalized to the $[0, 255]$ range.

For the coarse scene spatial extractor $\rm Extractor_{C}$, we propose a Transformer-based model, which has proven highly effective in capturing long-range dependencies~\cite{Swin}. Specifically, we propose the Deformable Window Super-Attention (DW-SA) Transformer block to enhance the window adaptation. Considering that deformable technology is empirically designed in the later stages of the network~\cite{dcn,dcn2}, the proposed DW-SA-T block replaces the Swin-T blocks in the third stage rather than the early stages. As shown in Figure \ref{fig4}, the DW-SA-T block introduces learnable offsets for each window. Since sub-pixel convolution was designed for SR image representation, each window is upsampled using sub-pixel convolution~\cite{subpixel} and then shifted following the protocol of Swin-T. The consecutive DW-SA Transformer blocks are computed as:
\begin{equation}
\small
\begin{aligned}
&\ \hat{\mathbf{z}}^{l}=\operatorname{W-MSA}\left(\operatorname{LN}\left(\mathbf{z}^{l-1}\right)\right)+\mathbf{z}^{l-1} \\
&\ \mathbf{z}^{l}=\operatorname{MLP}\left(\operatorname{LN}\left(\hat{\mathbf{z}}^{l}\right)\right)+\hat{\mathbf{z}}^{l} \\
&\ \hat{\mathbf{z}}^{l+1}=\operatorname{DW-SA}\left(\operatorname{LN}\left(\mathbf{z}^{l}\right)\right)+\mathbf{z}^{l} \\
&\ \mathbf{z}^{l+1}=\operatorname{MLP}\left(\operatorname{LN}\left(\hat{\mathbf{z}}^{l+1}\right)\right)+\hat{\mathbf{z}}^{l+1},
\end{aligned}
\end{equation}
where $\rm \hat{\mathbf{z}}^{l}$ and $\mathbf{z}^{l}$ denote the output features of the W-MSA and DW-SA modules and the MLP module for block $l$, respectively; W-MSA and LN denote window-based multi-head self-attention and layer norm of Swin-T, respectively.

For $\rm \hat{V_{D}}^{F}$, we utilize ResNet~\cite{resnet} as $\rm Extractor_{F}$ to capture spatial features. Both coarse and fine scene spatial extractors are pre-trained on ImageNet-1k~\cite{imagenet}. Given input $\rm \hat{V_{D}}^{C}, \hat{V_{D}}^{F} \in \mathbb{R}^{F\times W\times H\times 3}$, the coarse scene extractor first resizes $\rm \hat{V_{D}}^{C}$ from $\rm F\times W\times H\times 3$ to $\rm F\times 224\times224\times 3$, while the fine scene extractor processes $\rm \hat{V_{D}}^{F}$ directly. The final spatial features $\rm F_{S}\in \mathbb{R}^{F\times5632}$ are as follows:
\begin{equation}
\begin{aligned}
    &\ \rm F_{S}^{C} =\rm Extractor_{C}(\hat{V_{D}}^{C}), \rm F_{S}^{F} = Extractor_{F}(\hat{V_{D}}^{F}), \\
    &\ \rm F_{S} = Concatenate(V_{S}^{C},V_{S}^{F}).
\end{aligned}
\end{equation}
\subsection{Inconsistency Guided Temporal Module (IGTM)}
In this section, we propose a two-stage temporal aggregation integrating cross-time-scale relationships. 
\subsubsection{Consistency-aware Fusion stage}
One of the key factors in temporal aggregation is determining the appropriate amount of information that each time segment should load. Recent psychological research on the visual working memory mechanism~\cite{MC1,MC3,MC5} has yielded perceptual findings regarding memory capacity, showing that human visual working memory has a capacity limit when temporarily storing and processing visual information.

Inspired by these findings, we developed a \textbf{Visual Memory Capacity Block} that follows two principles:

1. Dynamic allocation of memory threshold in a range.

2. When the level of time inconsistency increases, the memory threshold decreases.

Principle 1 is derived from psychological research~\cite{MC1,MC2,MC5,MC6}, which suggests that the memory capacity of humans is approximately 3-7 objects and dynamically adjusted according to scene complexity. Principle 2 is supported by psychological studies~\cite{MC3,MC4}, which show that increased scene complexity negatively impacts memory performance.

We first quantify the temporal inconsistency levels. Since the video $\rm V_{I}$ reflects the level of temporally inconsistent changes in the SR video, we calculate the complexity of $\rm V_{I}$ to determine the memory capacity (i.e., threshold). Specifically, we first compute the spatial complexity of each frame and then obtain the overall complexity of the video. As $\rm V_{I}$ is based on optical flow, the magnitude of each frame and directional consistency are taken into account:
\begin{equation}
\small
\begin{aligned}
     \rm C_{I}^{ij} = \alpha\times(\sigma(M(V_{I}^{ij}))+(1-\alpha)\times(\sigma(D(V_{I}^{ij}))), \\
     \rm i=1,\dots,N; j =1,\dots,F.
\end{aligned}
\end{equation}
where $\rm N$ and $\rm F$ are the number of videos and frame count for each $\rm V_I$, respectively. $\rm M(\cdot)$ refers to magnitude computation, and $\rm D(\cdot)$ calculates the direction of the difference between adjacent frames, generating a histogram of the directions, and then the standard deviation $\sigma$ of the histogram represents the directional consistency. $\alpha$ is a hyperparameter. Then, the complexity of $\rm V_{I}$ is calculated as follows:
\begin{equation}
\small
\begin{aligned}
     \rm C_{I}^{i} = \mu(\{C_{I}^{ij}\|j=1,\dots,F\}) +\sigma(\{C_{I}^{ij}\|j=1,\dots,F\}), \\
     \rm i=1,\dots,N.
\end{aligned}
\end{equation}
where $\mu$ and $\sigma$ refer to mean and standard deviation, respectively. $\rm C_{I}=\{C_{I}^{i}\}$ is then normalized to $[0,1]$. The ablation analysis of $\alpha$ is displayed in Experiments. Then the input features $\rm F_{S}$ can be segmented and aggregated by the following visual memory capacity block, as shown in Algorithm~\ref{a1}.
\begin{algorithm}[!h]
\small    
\caption{Visual Memory Capacity Block}
    \renewcommand{\algorithmicrequire}{\textbf{Input:}}
    \renewcommand{\algorithmicensure}{\textbf{Output:}}
    \begin{algorithmic}[1]
        \REQUIRE Spatial features $\rm F_{S}^{i} \in \rm \mathbb{R}^{F\times 5632}$, temporal inconsistency levels $\rm C_{I}^{ij} \in \mathbb{R}^{F\times 1}$ and $\rm C_{I}^{i}\in \mathbb{R}^{1}$, where $\rm i=1,\dots,N;j=1,\dots,F$.
        \ENSURE Aggregated features of the first stage: $\rm F_{A}^{i}, i =1,\dots,N$.   
        \STATE Compute the adaptive memory threshold of SR video $\rm V_{D}^{i}$ as: \\
        $\rm T_{D}^{i} = \tau-\eta\times\frac{C_{I}^{i}-MIN(\{C_{I}^{i}\})}{MAX(\{C_{I}^{i}\})-MIN(\{C_{I}^{i}\})}$.
        \STATE$\#$  The adaptive $\rm T_{D}^{i}$ uses $\tau = 5$ and $\eta = 4$.
        \STATE Adaptively segment input features $\rm F_{S}^{i}$:
        \\Set of segments:$\rm S = \varnothing$;
        \\Current segment:$\rm S_{C} = \varnothing$;
        \FOR{$\rm j \in [1,F]$} 
            \STATE Current complexity:$\rm C_{C}+= C_{I}^{ij}$
            \STATE Add $\rm F_{S}^{ij}$ to $\rm S_{C}$: $\rm S_{C}.append(F_{S}^{ij})$
        \IF{$\rm C_{C} >= T_{D}^{i}$}  
                 \STATE $\rm S.append(S_{C})$
                 \STATE $\rm S_{C} = \varnothing$
                 \STATE $\rm C_{C} = 0$
        \ENDIF
        \ENDFOR
        \FOR{$\rm k \in N_{S}$}
        \STATE $\rm F_{A}^{ik} = Concatenate(MEAN(S_{k})+STD(S_{k}))$
        \ENDFOR
        \STATE $\#$Where $\rm N_{S}$ is the number of segments in $\rm S$.
    \end{algorithmic}
    \label{a1}
\end{algorithm}

Subsequently, we model the temporal relationships of the first-stage features $\rm F_{A} \in \{F_{A}^{i}\}$. Recognizing that Long Short-Term Memory models~\cite{lstm,GRU} are insufficient for capturing complex temporal dependencies, we propose the \textbf{Temporal Modeling} process as follows:

First, we calculate the sparse adjacency matrix of input features $\rm Adj(F_{A})$. Each node feature $\rm h_i$ of $\rm F_{A}$ is then transformed using a learnable weight matrix $\rm W$ to increase the representational capacity:
\begin{equation}
\small
 \rm W_h(i) = W h_i,
\end{equation}
where $\rm W$ is the weight matrix shared across all nodes. Then, the attention coefficients $\rm e_{ij}$ between node $\rm i$ and its neighbor $\rm j$ are computed by concatenating the transformed features of the two nodes, followed by applying a shared attention mechanism:
\begin{equation}
\small
\rm e_{ij} = \text{LeakyReLU}(a^T [W_h(i) \| W_h(j)]),
\end{equation}
where $\rm a$ is the learnable attention vector and $\|$ denotes concatenation. Then filter $\rm e=\{e_{ij}\}$ based on the positive elements of $\rm Adj(F_{A})$:
\begin{footnotesize}
$$\rm e_{ij}=
\begin{cases}
\rm e_{ij}& \rm Adj(F_{A})_{ij}>0\\
\rm 0& \rm Adj(F_{A})_{ij}\leq 0
\end{cases}$$
\end{footnotesize}
The attention coefficients are normalized using the softmax function:
\begin{equation}
\small
\rm
\alpha_{ij} = \frac{\exp(e_{ij})}{\sum_{k \in \mathcal{N}(i)} \exp(e_{ik})},
\end{equation}
where $\rm \mathcal{N}(i)$ represents the neighbors of node $\rm i$. Finally, the new feature $\rm h'_i$ of node $\rm i$ is computed by aggregating its neighbors’ features weighted by the attention coefficients:
\begin{equation}
\small
\rm
 h'_i = \phi \left( \sum_{j \in \mathcal{N}(i)} \alpha_{ij} W_h(j) \right),
\end{equation}
where $\rm \phi$ is a non-linear activation function. And the features after time modeling are obtained by the Gated Recurrent Unit (GRU)~\cite{GRU}:
\begin{equation}
\small
\rm
F_{S_1} = \text{GRU}(\{h'_{i}\}).
\end{equation}
\begin{table*}[t]
\centering
\tiny
  \caption{Performance comparison of our methods against competing IQA/VQA methods on both Single-Frame SR videos and Multi-Frame
SR videos. The best and second-best performances are highlighted in \textbf{bold} and \underline{underlined}, respectively.}
  \resizebox{1.0\linewidth}{!}{
\begin{tabular}{c|cccc|cccc|cccc}
\hline
Datasets & \multicolumn{4}{c|}{SFD} & \multicolumn{4}{c|}{MFD}  & \multicolumn{4}{c}{Combined-VSR}\\ \hline
\multicolumn{1}{c|}{Methods} & SRCC$\uparrow$ & PLCC$\uparrow$ & KRCC$\uparrow$ & RMSE$\downarrow$ & SRCC$\uparrow$ & PLCC$\uparrow$ & KRCC$\uparrow$ & RMSE$\downarrow$ & SRCC$\uparrow$ & PLCC$\uparrow$ & KRCC$\uparrow$ & RMSE$\downarrow$ \\ \hline
  PSNR & 0.654 & 0.661 & 0.476 & 0.197 & 0.639 & 0.648 & 0.459 & 0.202 & 0.645 & 0.655 & 0.468 & 0.200\\
  SSIM~\cite{SSIM} & 0.709 & 0.719 & 0.534 & 0.185 & 0.693 & 0.701 & 0.519 & 0.190 & 0.696 & 0.710 & 0.525 & 0.189 \\
  VIF~\cite{VIF} & 0.746 & 0.755 & 0.579 & 0.164 & 0.742 & 0.749 & 0.571 & 0.166 & 0.746 & 0.753 & 0.579 & 0.165\\
  SpEED-QA~\cite{SpEED-QA} & 0.507 & 0.520 & 0.309 & 0.243 & 0.501 & 0.514 & 0.303 & 0.247 & 0.504 & 0.516 & 0.307 & 0.245\\
  IGTS~\cite{IGTS} & 0.535 & 0.552 & 0.338 & 0.232 & 0.530 & 0.547 & 0.334 & 0.235 & 0.533 & 0.550 & 0.337 & 0.234\\
  VSFA~\cite{VSFA} & 0.813 & 0.819 & 0.636 & 0.150 & 0.796 & 0.801 & 0.619 & 0.158 & 0.808 & 0.812 & 0.630 & 0.152 \\
  DeepSRQ~\cite{deepsrq} & 0.671 & 0.661 & 0.480 & 0.192 & 0.651 & 0.643 & 0.462 & 0.201 & 0.667 & 0.656 & 0.474 & 0.195\\
  VMAF~\cite{VMAF} & 0.712 & 0.729 & 0.545 & 0.182 & 0.706 & 0.713 & 0.528 & 0.188 & 0.710 & 0.723 & 0.539 & 0.184\\
  DR-IQA~\cite{DR-IQA} & 0.717 & 0.723 & 0.542 & 0.179 & 0.701 & 0.711 & 0.532 & 0.188 & 0.707 & 0.716 & 0.537 & 0.184 \\
  VIDEVAL~\cite{VIDEVAL} & 0.754 & 0.759 & 0.583 & 0.164 & 0.731 & 0.740 & 0.569 & 0.171 & 0.744 & 0.749 & 0.577 & 0.167 \\
  DISQ~\cite{DISQ} & 0.665 & 0.661 & 0.475 & 0.198 & 0.636 & 0.641 & 0.459 & 0.207 & 0.642 & 0.650 & 0.465 & 0.202 \\
  SRIF~\cite{SRIF} & 0.751 & 0.758 & 0.582 & 0.167 & 0.739 & 0.744 & 0.570 & 0.171 & 0.743 & 0.751 & 0.576 & 0.169 \\
  GSTVQA~\cite{GSTVQA} & 0.839 & 0.837 & 0.655 & 0.141 &  0.816 & 0.815 & 0.633 & 0.153 & 0.828 & 0.825 & 0.645 & 0.147 \\
  STF~\cite{STF} & 0.776 & 0.783 & 0.634 & 0.158 & 0.754 & 0.758 & 0.583 & 0.169 & 0.763 & 0.769 & 0.594 & 0.163 \\
  STI-VQA~\cite{STIVQA} & 0.835 & 0.842 & 0.658 & 0.139 & 0.814 & 0.819 & 0.636 & 0.152 & 0.823 & 0.829 & 0.648 & 0.147 \\
  2Bi-VQA~\cite{2BIVQA} & 0.820 & 0.838 & 0.628 & 0.125 & 0.774 & 0.809 & 0.569 & 0.140 & 0.777 & 0.810 & 0.585 & 0.665 \\
 FAST-VQA~\cite{fastvqa} & 0.852 & 0.861 & 0.661 & 0.130 & 0.836 & 0.851 & 0.639 & 0.134 & 0.845 & 0.856 & 0.651 & 0.132 \\
  MBVQA~\cite{MBVQA} & \underline{0.892} & \underline{0.901} & \underline{0.717} & \underline{0.106} & 0.795 & 0.827 & 0.606 & 0.124 & 0.840 & 0.853 & 0.644 & 0.127 \\ 
  VSR-QAD\cite{VSRQAD} & 0.884 & 0.889 & 0.706 & 0.115 & 0.846 & 0.846 & 0.672 & 0.134 & 0.860 & 0.868 & 0.687 & 0.125 \\
  ReLaX-VQA~\cite{ReLaX-VQA} & \underline{0.937} & \underline{0.949} & \underline{0.785} & \underline{0.100} & \underline{0.915} & \underline{0.925} & \underline{0.747} & \underline{0.098} & \underline{0.924} & \underline{0.936} & \underline{0.782} & \underline{0.091} \\
  \textbf{TIG-SVQA} & \textbf{0.950} & \textbf{0.951} & \textbf{0.803} & \textbf{0.093} & \textbf{0.942} & \textbf{0.943} & \textbf{0.790} & \textbf{0.091} & \textbf{0.939} & \textbf{0.942} & \textbf{0.794} & \textbf{0.083} \\ \hline
\end{tabular}}
\label{t1}
\end{table*}
\begin{table}[t]
\centering
\tiny
\caption{Ablation experiments on each component of TIG-SVQA on the Combined-VSR dataset. Note that “w/o Guidance in IGTM” is to replace the visual memory capacity block with fixed temporal segments.}
\resizebox{0.85\linewidth}{!}{
\begin{tabular}{c|cccc}
\hline
Methods & SRCC$\uparrow$ & PLCC$\uparrow$ & KRCC$\uparrow$ & RMSE$\downarrow$ \\ \hline
w/o Guidance in IHSM & 0.891 & 0.909 & 0.716 & 0.116 \\
w/o Guidance in IGTM  & 0.908 & 0.921 & 0.736 & 0.095 \\
w/o both Guidance & 0.878 & 0.901 & 0.707 & 0.107\\
Coarse Branch only in IHSM & 0.789 & 0.846 & 0.609 & 0.131 \\
Fine Branch only in IHSM & 0.926 & 0.927 & 0.771 & 0.106 \\
w/o DW-SA-T block & 0.891 & 0.909 & 0.716 & 0.116 \\
\textbf{Proposed TIG-SVQA} & \textbf{0.939} & \textbf{0.942} & \textbf{0.794} & \textbf{0.083} \\ \hline
\end{tabular}}
\label{t2}
\end{table}

These processes together make up one temporal modeling process. $\rm F_{S_1}$ is then regressed to score $\rm S_1$, and used for the second-stage temporal aggregation. 
\subsubsection{Informative Filtering stage}
In this stage, we filter out the most informative features to facilitate temporal relationship modeling. Specifically, self-attention is computed over $\rm F_{S_1}$, and the top $\rm K$ features with the highest attention scores are selected as key features:
\begin{equation}
\small
\rm W_S = SA(F_{S_1}), F_{S_2} = F_{S_1}[Top_{K}(W_S)],
\end{equation}
where $\rm SA$ represents self-attention. $\rm W_S$ is the attention weight. The $\rm F_{S_2}$ is then processed by another temporal modeling process, which regresses the second-stage quality score $\rm S_{2}$. The final quality prediction is computed as:
\begin{equation}
\small
\rm S = \gamma\times S_{1}+(1-\gamma)\times S_{2},
\end{equation}
where $\gamma$ is a hyperparameter.
\section{Experiments}
\subsection{Experimental Setups}
\textbf{Benchmark Datasets}. We evaluate TIG-SVQA on both single-frame and multi-frame super-resolution video quality datasets, i.e., SFD and MFD. The SFD~\cite{VSRQAD} contains 1,193 videos generated from 120 reference sequences, downsampled by factors of ×2, ×4, and ×8, and then upscaled using five different single-frame SR methods. In contrast, the MFD~\cite{VSRQAD} comprises 1,067 videos produced by five state-of-the-art multi-frame SR algorithms. The Combined-VSR dataset includes all 2,260 videos, spanning 10 recent SR methods across both single-frame and multi-frame categories.
\begin{table}[t]
\small
\centering
  \caption{Comparisons regarding model complexity. Note that the number of parameters of methods that contain several modules is the sum of each module. The best and second-best performances are highlighted in \textbf{bold} in \underline{underlined}, respectively.}
   \resizebox{0.8\linewidth}{!}{
\begin{tabular}{c|ccc}
\hline
Model   & Flops/G $\downarrow$ & Paras/M $\downarrow$ & SRCC$\uparrow$ \\ \hline
DISQ~\cite{DISQ}  & 606.69 & 76.18 & 0.642 \\
FAST-VQA~\cite{fastvqa} & \textbf{70.90} & 27.55 & 0.845\\
STI-VQA~\cite{STIVQA} & 103087.70 & 89.37 & 0.823 \\
MBVQA~\cite{MBVQA}   & 2149.90 & 93.23 & 0.840\\
VSR-QAD~\cite{VSRQAD} & 678.95 & \textbf{23.74} & \underline{0.860} \\
\textbf{Proposed TIG-SVQA} & \underline{171.63} & \underline{24.96} & \textbf{0.939}\\ \hline
\end{tabular}}
\label{t3}
\end{table}

\textbf{Competing Methods}. We compare our method with 8 image quality assessment (IQA) \cite{SSIM,VIF,SRIF,IGTS,DISQ,DR-IQA,STF,deepsrq}, and 10 video quality assessment (VQA) methods~\cite{VIDEVAL,VMAF,SpEED-QA,VSRQAD,VSFA,GSTVQA,STIVQA,2BIVQA,fastvqa,MBVQA}. Among them, SRIF, IGTS, DISQ, STF, DeepSRQ are SR IQA methods, and VSR-QAD is the latest SR VQA method.

\textbf{Performance Criteria}. We evaluate performance using Spearman rank-order correlation coefficient (SRCC), Kendall rank-order correlation coefficient (KRCC), Pearson linear correlation coefficient (PLCC), and root mean square error (RMSE). For each dataset, we follow the protocols of ~\cite{VSRQAD}: randomly splitting videos into $70\%$ training, $10\%$ validation, and $20\%$ testing.

\textbf{Implementation Details}. Models were trained for 100 epochs on a 16GB NVIDIA RTX 3080 Ti with PyTorch 1.7.1. The Adam optimizer was used with an initial learning rate of $10^{-5}$, decaying by 0.8 every 10 epochs. The batch size was 16, with no weight decay. The loss combined SRCC and MSE to leverage both ranking and regression.

\subsection{Performance on SFD, MFD, and Combined-VSR}
Table \ref{t1} presents a comprehensive comparison of the latest IQA/VQA methods on single-frame and multi-frame SR video quality datasets. Among the listed methods, PSNR and SSIM are classical handcrafted metrics, relying on pixel-wise differences without learning, while VIF and SpEED-QA are also handcrafted metrics utilizing signal fidelity and statistical priors. In contrast, methods such as DeepSRQ, DR-IQA, FAST-VQA, 2Bi-VQA, MBVQA, and VSR-QAD are learning-based approaches. Notably, SRIF, IGTS, DISQ, STF, and DeepSRQ are originally proposed for SR image quality assessment, whereas VSR-QAD is the latest VQA method designed specifically for SR videos.

The performance comparison reveals that the proposed TIG-SVQA significantly outperforms both handcrafted and learning-based baselines across all settings. Compared to handcrafted metrics like VIF and SpEED-QA, TIG-SVQA achieves substantial improvements in correlation metrics and error reduction, particularly on the more challenging MFD subset. Even when compared to traditional VQA methods such as FAST-VQA and MBVQA, or the latest SR VQA method, i.e., VSR-QAD, TIG-SVQA shows superior results. These findings highlight the necessity for dedicated temporal inconsistency guidance for learning SR video distortions.

\subsection{Ablation Study}
\label{lala}
The ablation of each component in TIG-SVQA highlights the individual contributions, as shown in Table \ref{t2}. \textbf{w/o Guidance in IHSM} refers to using SR videos as input without weighting with temporal inconsistency information in the IHSM. \textbf{w/o Guidance in IGTM} refers to replacing the visual memory capacity block in the temporal module with fixed temporal segments (i.e., one batch has 16 frames). \textbf{w/o both Guidance} refers to not using temporal inconsistency guidance in both IHSM and IGTM. \textbf{Coarse/Fine Branch only in IHSM} refers to only using the coarse/fine scene spatial extractor. And \textbf{w/o DW-SA-T block} refers to utilizing Swin-T blocks rather than the proposed DW-SA-T blocks in the IHSM. The model's performance drops notably without the DW-SA Transformer blocks and inconsistency guidance, highlighting their critical role in perceptual prediction. The coarse/fine spatial branch alone shows a marked decline, emphasizing the need to combine both granularity features.
\begin{table}[t]
\centering
 \caption{Validation of memory thresholds in the visual memory capacity block. Note that the equation for $\rm T_D^{i}$ in Algorithm 1 supports adaptive processing.}
\resizebox{0.8\linewidth}{!}{
\small
\begin{tabular}{c|cccc}
\hline
Memory Thresholds & SRCC$\uparrow$ & PLCC$\uparrow$ & KRCC$\uparrow$ & RMSE$\downarrow$ \\ \hline
1 & 0.931 & 0.935 & 0.772 & 0.095 \\
5 & 0.925 & 0.932 & 0.770 & 0.104 \\
10 & 0.912 & 0.918 & 0.750 & 0.102 \\
15 & 0.895 & 0.905 & 0.727 & 0.104 \\
Adaptive 1 to 5 & \textbf{0.939} & \textbf{0.942} & \textbf{0.794} & \textbf{0.083} \\
Adaptive 1 to 10 & 0.933 & 0.936 & 0.782 & 0.083 \\ \hline
\end{tabular}}
\label{t4}
\end{table}
\subsection{Model Complexity Analysis and Parameter Test}
\label{ablation}
Table \ref{t3} compares state-of-the-art VQA methods in terms of \textbf{FLOPs and parameter count}. FLOPS is shown in GigaFLOPS (G), while the parameter count is shown in Mbyte (M). We assume that the resolution of the test video is 1080P, the duration is 6 seconds, and the number of video frames is 150 frames. It can be observed that TIG-SVQA has the best correlation performance and the second-best regarding FLOPs as well as parameters.

Table \ref{t4} shows that increasing \textbf{memory thresholds} from 1 to 15 decreases the performance (e.g., SRCC decreases from 0.931 to 0.895). However, using adaptive memory thresholds generally improves the final performance. This suggests that dynamic memory allocation enhances the model's accuracy in video quality prediction, which aligns better with the human visual system. Table \ref{t5} presents an ablation on \textbf{hyperparameter $\alpha$}. The model performs best at $\alpha = 0.5$. Performance decreases at the extremes ($\alpha$ = 0.1 and $\alpha$ = 0.9), indicating that a balanced approach between the two complexities is the most effective. The \textbf{hyperparameter $\gamma$} is used to weight the prediction quality scores $S_1$ and $S_2$ of the two stages. As shown in Table \ref{t6}, as $\gamma$ gradually increases from 0 to 1, the performance first increases and then decreases, with the best performance in the range of $[0.4,0.6]$.

\begin{table}[t]
  \centering
  \small
  \caption{Ablation on the hyper-parameter $\alpha$, which determines the weight of the magnitude complexity and complexity of direction consistency when computing the level of temporal inconsistency for each frame.}
  \resizebox{0.8\linewidth}{!}{
    \begin{tabular}{c|cccc}
    \hline
    Hyper-Param-$\alpha$ & SRCC$\uparrow$ & PLCC$\uparrow$ & KRCC$\uparrow$  & RMSE$\downarrow$ \\
    \hline
    0.1   & 0.931 & 0.935 & 0.782 & 0.101 \\
    0.3   & 0.933 & 0.941 & 0.779 & 0.100 \\
    0.5   & \textbf{0.939} & \textbf{0.942} &\textbf{0.794}  & \textbf{0.083} \\
    0.7   & 0.930 & 0.935 & 0.784 & 0.087 \\
    0.9   & 0.929 & 0.936 & 0.782 & 0.103 \\
    \hline
    \end{tabular}}
  \label{t5}%
\end{table}%
\begin{table}[t]
  \centering
 \small
  \caption{Ablation on the hyper-parameter $\gamma$, which determines the proportion of predicted scores $S_{1}$ and $S_{2}$ at cross-time-scales in the final predicted score $S$.}
  \resizebox{0.8\linewidth}{!}{
    \begin{tabular}{c|cccc}
    \hline
     Hyper-Param-$\gamma$ & SRCC$\uparrow$  & PLCC$\uparrow$  & KRCC$\uparrow$  & RMSE$\downarrow$ \\
    \hline
    0.0   & 0.919 & 0.931 & 0.764 & 0.106 \\
    0.2   & 0.926 & 0.935 & 0.773 & 0.094 \\
    0.4   & 0.932 & 0.938 & 0.783 & 0.091 \\
    0.5   & \textbf{0.939} & \textbf{0.942} & \textbf{0.794}  & \textbf{0.083} \\
    0.6   & 0.928 & 0.936 & 0.779 & 0.090 \\
    0.8   &  0.918 & 0.927 & 0.764 & 0.090  \\
    1.0   & 0.910 & 0.920 & 0.748 & 0.092 \\
    \hline
    \end{tabular}}
  \label{t6}%
\end{table}%
\begin{figure*}[t] \centering
    \includegraphics[width=\textwidth]{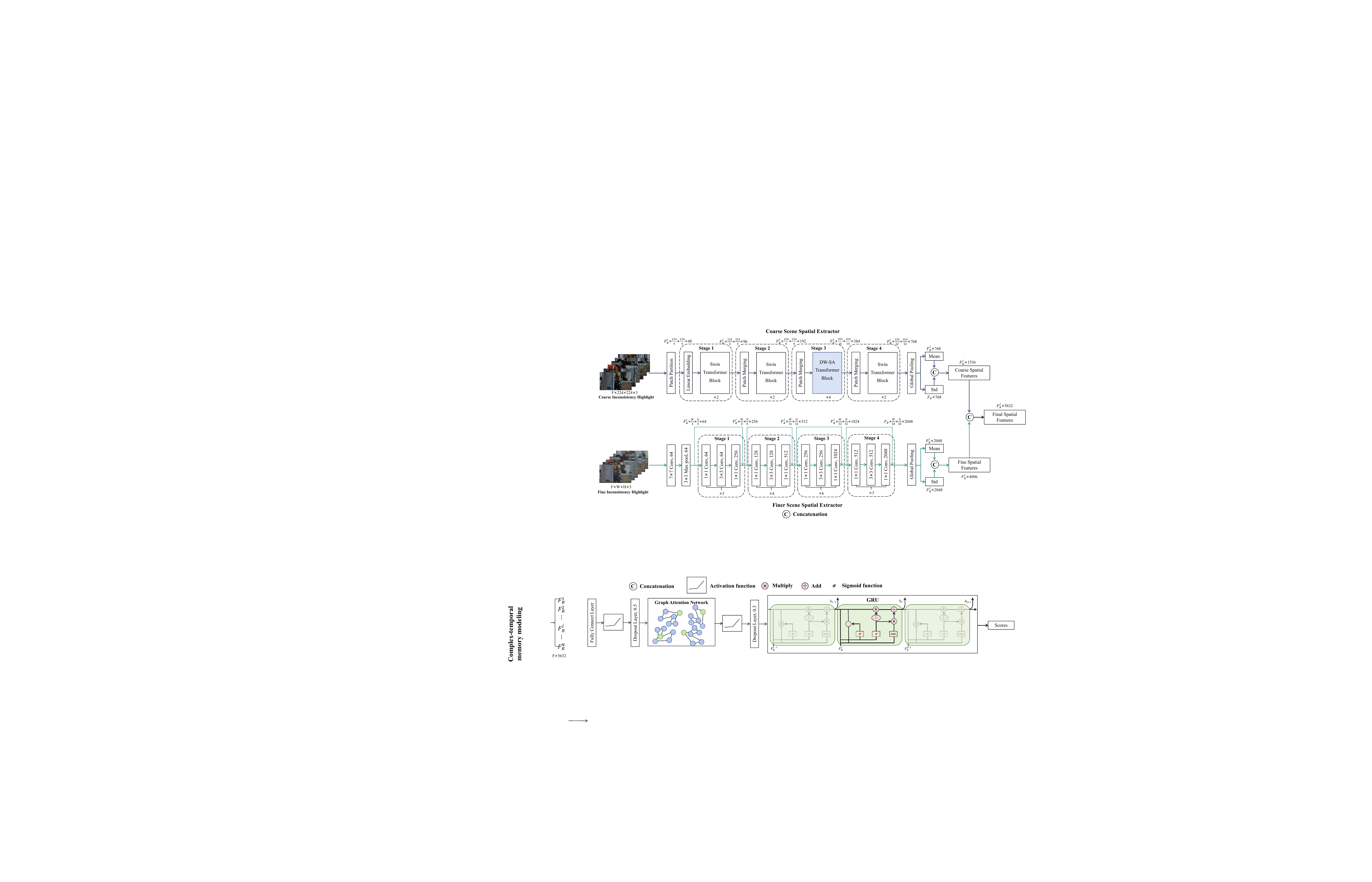}
    \caption{The framework of \textbf{Inconsistency Highlighted Spatial Module (IHSM)} includes detailed structures for both the Coarse Scene Spatial Extractor and the Fine Scene Spatial Extractor. Each layer of these extractors, along with the input and output feature dimensions, is described. The input frame batch for the $j$-th iteration is denoted as $F_{B}^{j}$, where $B$ represents the batch size, and $W$ and $H$ correspond to the frame's width and height, respectively.} 
\label{fig:IHSMspatial}
\end{figure*}
\begin{figure*}[t] \centering
    \includegraphics[width=\textwidth]{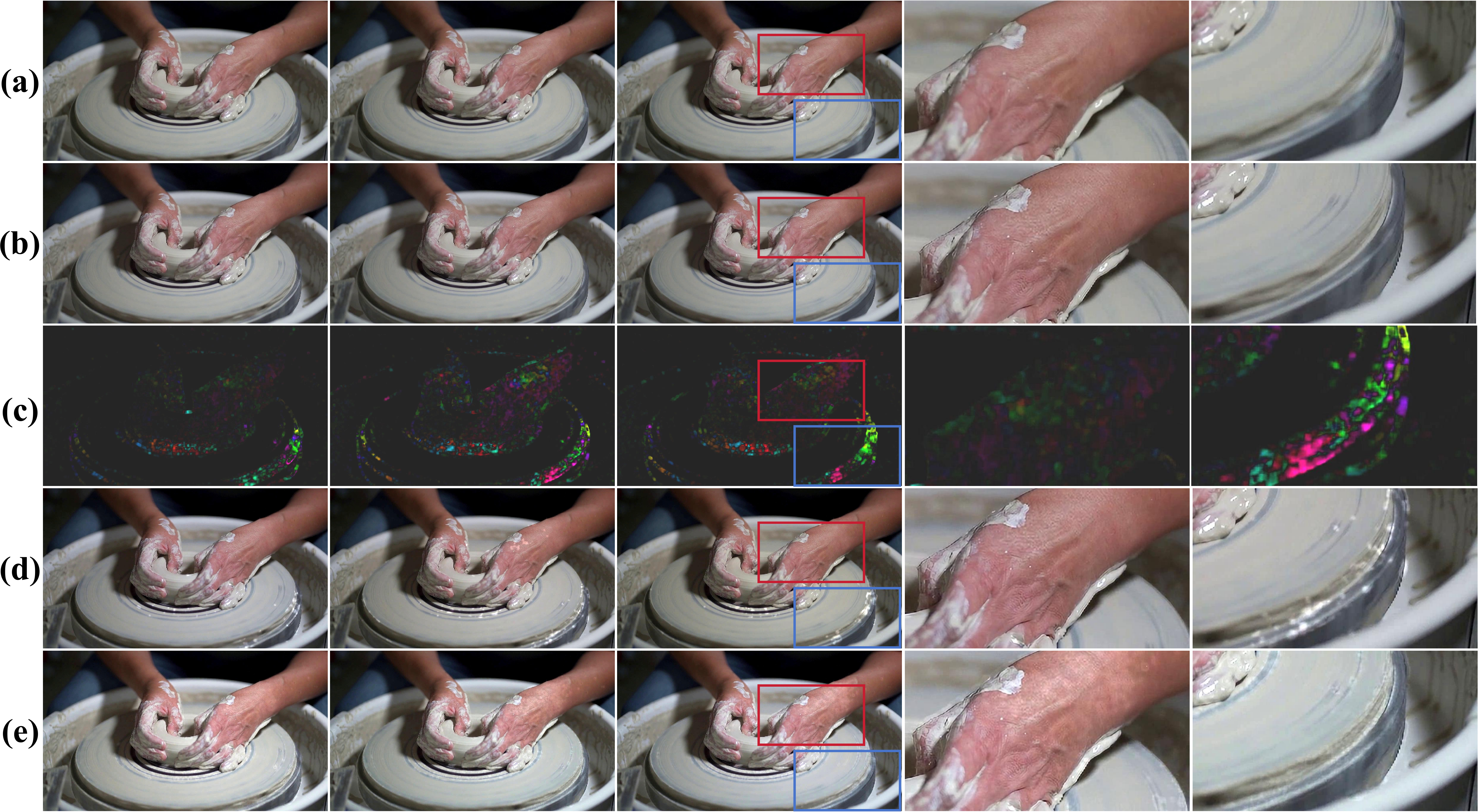}
    \caption{\textbf{Visualization of the highlighted regions in IHSM.} (a) shows the reference video, and (b) is the super-resolved (SR) video with a $\times4$ upsampling scale. (c) presents the computed temporal inconsistency map. (d) and (e) display the coarse and fine inconsistency highlighted results (highlighted in white), respectively. For clarity, the regions marked by red and blue boxes are enlarged and shown in the last two columns.} 
\label{fig:2}
\end{figure*}
\begin{figure}[t]
    \begin{minipage}[b]{0.45\linewidth}
        \includegraphics[width=\linewidth]{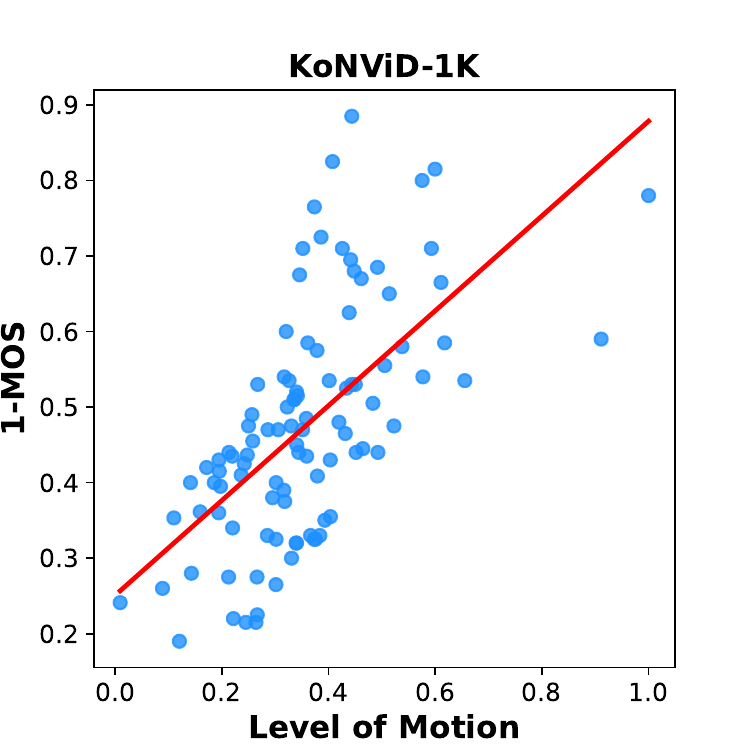} 
    \end{minipage}
    \hfill
    \begin{minipage}[b]{0.45\linewidth}
        \includegraphics[width=\linewidth]{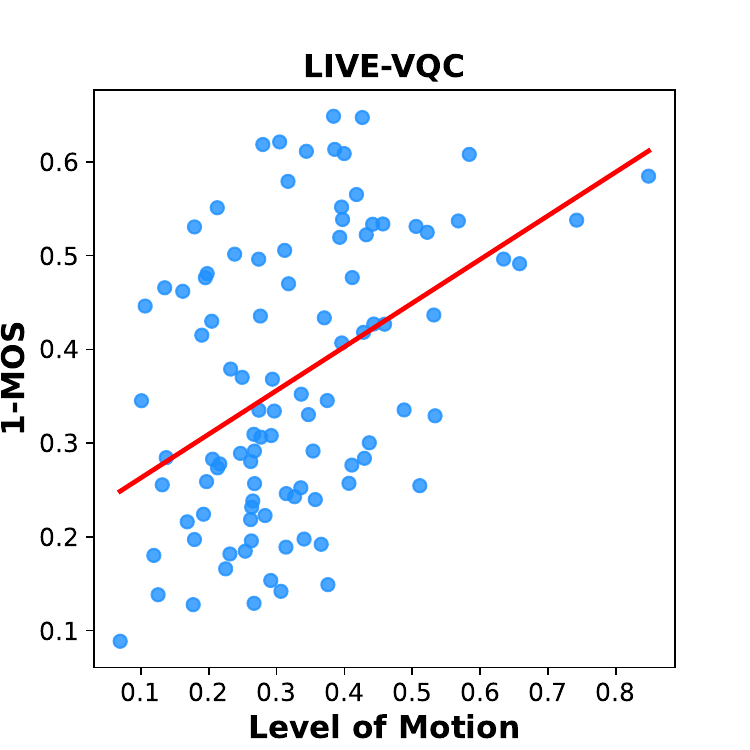} 
    \end{minipage}

    \caption{\textit{Empirical support for directly employing the motion information as the temporal inconsistency signal for UGC scenarios.} To investigate the critical correlation between motion level and perceptual quality, a random sample of 100 videos from each UGC dataset is analyzed. The results reveal an obvious correlation between the motion level and 1-MOS for UGC videos.}
    \label{fig:6}
\end{figure}
\section{More Details for the IHSM}
The Inconsistency Highlighted Spatial Module (IHSM) in the proposed TIG-SVQA method is detailed in Figure~\ref{fig:1}. This module is designed to capture two distinct types of temporal inconsistency:
\begin{itemize}
\item Coarse inconsistency highlight targets artifacts arising from large scene transitions and rapid motion, which require effective modeling of global dependencies.
\item Fine inconsistency highlight, on the other hand, addresses subtle distortions caused by slow motion, emphasizing the importance of fine-grained local feature extraction.
\end{itemize}
To this end, the IHSM integrates two specialized spatial extractors:

1) Coarse Scene Spatial Extractor

We introduce the Deformable Window Super Attention (DW-SA) Transformer as the coarse extractor, which replaces the third stage of the standard Swin Transformer with DW-SA blocks. This design is grounded in the following theoretical considerations:
\begin{itemize}
\item First, deformable convolutions have demonstrated notable effectiveness in dense prediction tasks, including image quality assessment (IQA)~\cite{radn,dqm-iqa}. Since Swin-T employs fixed sampling locations for windows, thus, incorporating learnable offsets to dynamically adapt window positions may offer additional performance gains.
\item Second, prior studies~\cite{dcn,dcn2} suggest that applying deformable mechanisms in later stages of the network yields better results. In contrast, early-stage deployment may introduce optimization instability or convergence challenges.
\end{itemize}
\begin{table}[t]
	\centering
	\caption{Ablation study on the placement of the DW-SA Transformer block within the IHSM on the Combined-VSR dataset. 
	In each experiment, Swin Transformer blocks in different stages are progressively replaced with DW-SA Transformer blocks to examine the impact of their position on model performance. 
	To exclusively evaluate the effect of this branch, all experiments use only the scene-level spatial feature extractor as an independent branch for validation. The best performance is highlighted in bold.}
	\label{tab:dwsat}
    \resizebox{1.0\linewidth}{!}{
	\begin{tabular}{c|cccc}
		\hline
		Position of DW-SA Transformer & SRCC$\uparrow$ & PLCC$\uparrow$ & KRCC$\uparrow$ & RMSE$\downarrow$ \\
		\hline
		Stage 1 & 0.784 & 0.865 & 0.604 & 0.124 \\
		Stage 2 & 0.784 & 0.864 & 0.604 & 0.119 \\
		Stage 3 & \textbf{0.825} & \textbf{0.886} & \textbf{0.646} & \textbf{0.113} \\
		Stage 4 & 0.811 & 0.879 & 0.628 & 0.117 \\
		\hline
	\end{tabular}}
\end{table}

Table \ref{tab:dwsat} presents the ablation results of inserting the DW-SA Transformer block at different positions within the scene-level spatial feature extractor. Experimental results demonstrate that the insertion position of this module has a significant impact on model performance. Specifically, introducing it at Stage 3 achieves the best performance (SRCC = 0.825, PLCC = 0.886), which is notably superior to other configurations.

A detailed analysis reveals the following performance ranking across stages: Stage 3 > Stage 4 > Stage 1 $\approx$ Stage 2. In terms of SRCC, Stage 3 outperforms Stage 4 (0.811) and Stage 1/2 (0.784) by 1.73$\%$ and 5.74$\%$, respectively. This observation aligns well with the theoretical characteristics of deformable convolutions: in deeper network stages (e.g., Stage 3), feature maps contain richer semantic information but lower spatial resolution. Introducing a deformable self-attention mechanism at this depth allows more effective modeling of geometric transformations, thereby enhancing the ability to capture inconsistencies caused by large-scale motion.

In contrast, inserting the DW-SA Transformer in shallower stages (Stage 1/2) yields limited improvement, likely because early-stage features mainly encode low-level texture information, which can already be effectively captured by fixed local windows. The performance drop observed in the final stage (Stage 4) may be attributed to its smaller depth (only two layers), whereas Stage 3 contains six layers (see Figure \ref{fig:IHSMspatial}). These findings provide strong empirical evidence for determining the optimal placement of the deformable wide self-attention module in super-resolution video quality assessment tasks.
2) Fine Scene Spatial Extractor

We adopt a standard ResNet~\cite{resnet} pretrained on the ImageNet-1K dataset to serve as the fine extractor, providing strong local feature representations with proven generalization capability.

By jointly leveraging both global and local spatial features through the coarse and fine extractors, respectively, TIG-SVQA achieves superior performance in capturing cross-granularity inconsistencies.

To illustrate the distinction between coarse and fine highlights, Figure \ref{fig:2} visualizes the decoupled inconsistency information at different granularities. As shown, distortions caused by rapid motion (e.g., the edge of a spinning disk) are highlighted as coarse inconsistency, while distortions from slower motion (e.g., the skin on a moving hand) are highlighted as fine inconsistency.
\begin{figure*}[h] \centering
    \includegraphics[width=\textwidth]{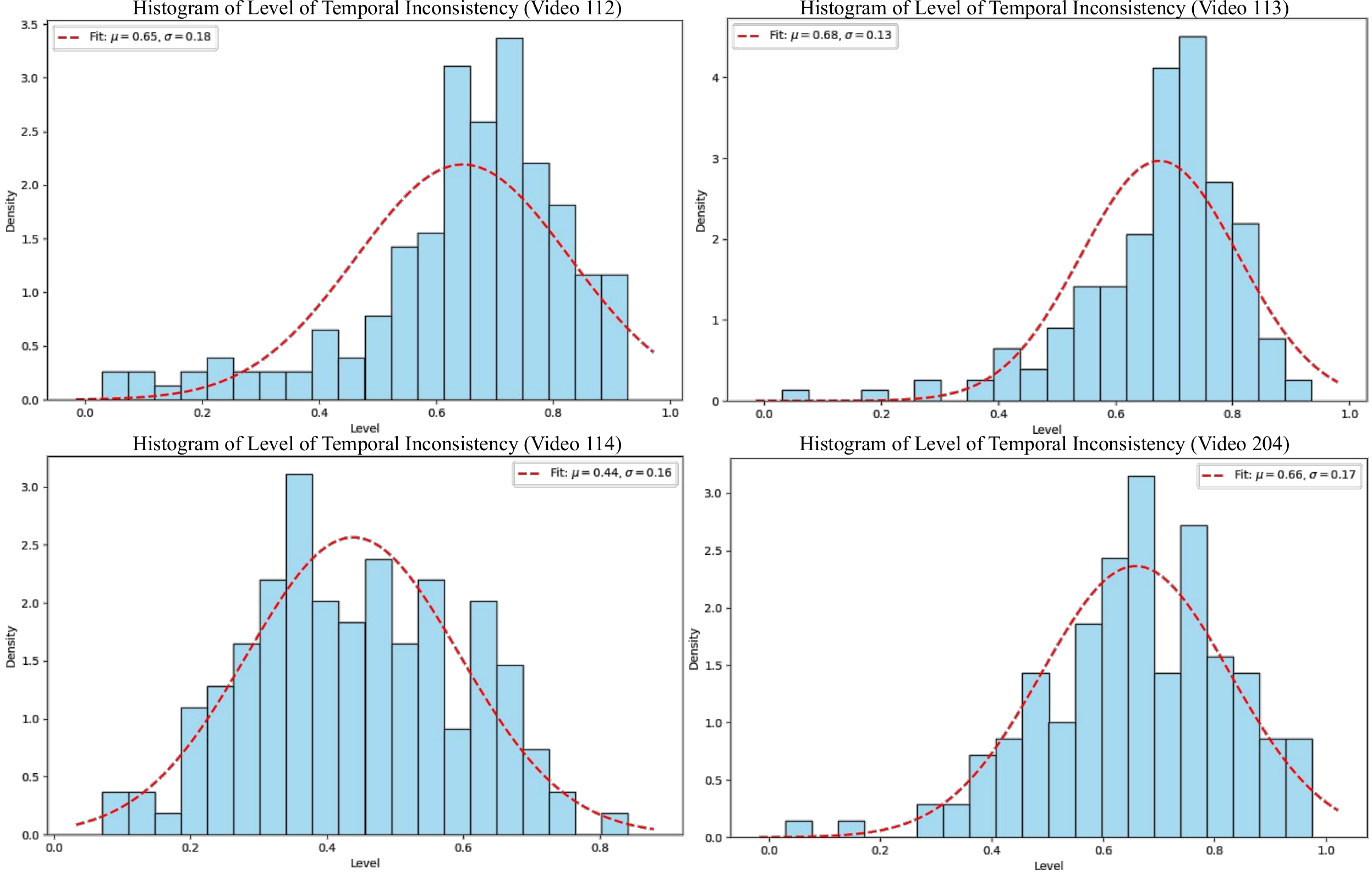}
    \caption{Examples of the histograms of the distribution of temporal inconsistency level in all frames of four videos. It can be observed that the distributions can be fitted as the Gaussian distribution.} \label{fig:3}
\end{figure*}
\begin{figure*}[t] \centering
    \includegraphics[width=\textwidth]{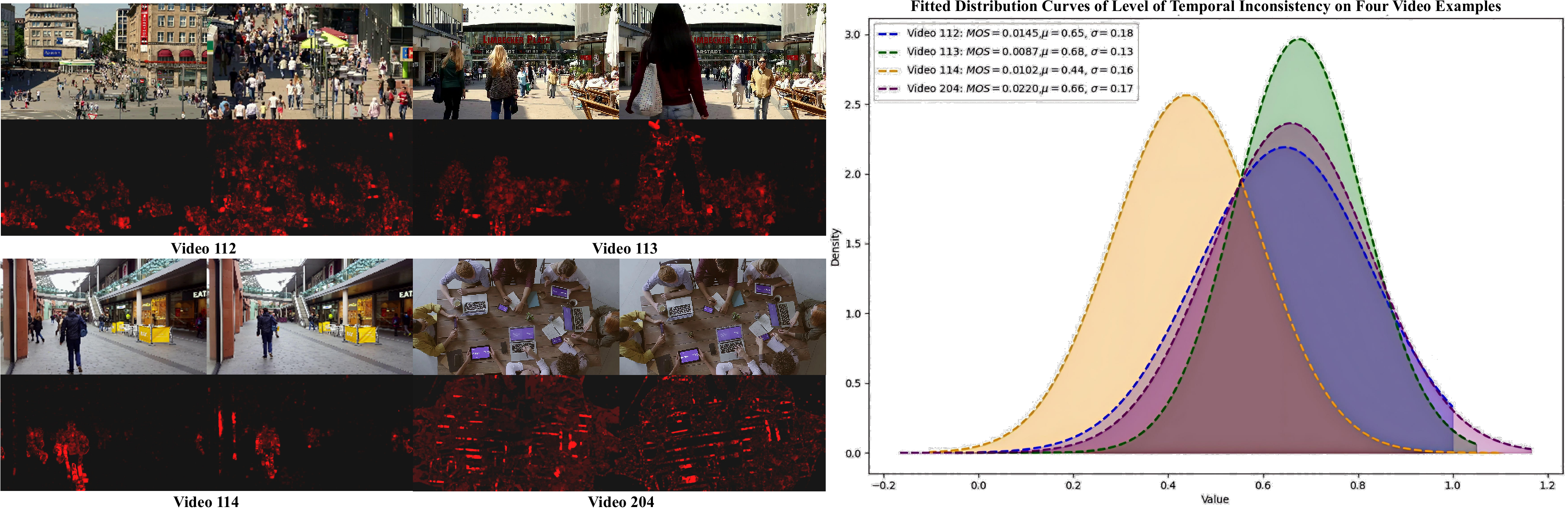}
    \caption{Four examples about the distributions of level of temporal inconsistency and sampled frames. It can be observed that the distributions of SR videos are suitable for Gaussian distribution, and videos have different levels of temporal inconsistency; thus, we design an adaptively allocated memory threshold in Eq. \ref{eq1}.} \label{fig:4}
\end{figure*}
\section{More Details for the IGTM}
In this section, we analyze the rationale behind the adaptive memory threshold design in the Visual Capacity Limit block of the Inconsistency Guided Temporal Module (IGTM). Specifically, the threshold $\rm T_D^{i}$ for each video $\rm V_I$ is determined based on its overall temporal inconsistency level $\rm C_I^{i}$, calculated as:
\begin{equation}
\small
\begin{aligned}
\rm C_{I}^{ij} &= \rm \alpha \cdot \sigma(M(V_{I}^{ij})) + (1 - \alpha) \cdot \sigma(D(V_{I}^{ij})), \\
\rm C_{I}^{i} &= \rm \mu\left(\{C_{I}^{ij} \mid j = 1, \dots, F\}\right) + \sigma\left(\{C_{I}^{ij} \mid j = 1, \dots, F\}\right), \\
& \rm \text{for } j = 1,\dots,F,
\end{aligned}  
\end{equation}
where $\rm F$ denotes the number of frames per video. Each frame-level inconsistency $\rm C_{I}^{ij}$ is derived from a weighted combination of the magnitude and directional consistency metrics, normalized via standard deviation $\rm \sigma(\cdot)$. The overall video-level inconsistency $\rm C_{I}^{i}$ is estimated by the sum of the mean $\rm \mu(\cdot)$ and standard deviation $\rm \sigma(\cdot)$ of the frame-level values.

To justify this formulation, we randomly selected 100 videos from the Combined-VSR dataset and calculated the histograms of their frame-level inconsistency values $\rm \{C_{I}^{ij}\}$. Figure~\ref{fig:3} displays 4 visualization examples of these histograms. These distributions closely approximate Gaussian curves, indicating that the combination $\mu + \sigma$ is a statistically reasonable indicator of the global inconsistency level $\rm C_I^i$.

As illustrated in Figure~\ref{fig:4}, different videos exhibit diverse inconsistency levels due to varying distortion severities. To accommodate this, we dynamically assign the memory threshold using min-max normalization:
\begin{equation}
\begin{aligned}
\rm T_{D}^{i} = 5 - 4 \cdot \frac{C_{I}^{i} - \min(\{C_{I}^{i}\})}{\max(\{C_{I}^{i}\}) - \min(\{C_{I}^{i}\})}, \\
\rm \text{for } i = 1,\dots,N,
\end{aligned}
\label{eq1}
\end{equation}
where $\rm N$ refers to the number of videos. This ensures that videos with higher temporal inconsistency are allocated smaller thresholds, promoting finer temporal segmentation and more adaptive memory control.
\begin{figure*}[t] 
\centering
\includegraphics[width=\textwidth]{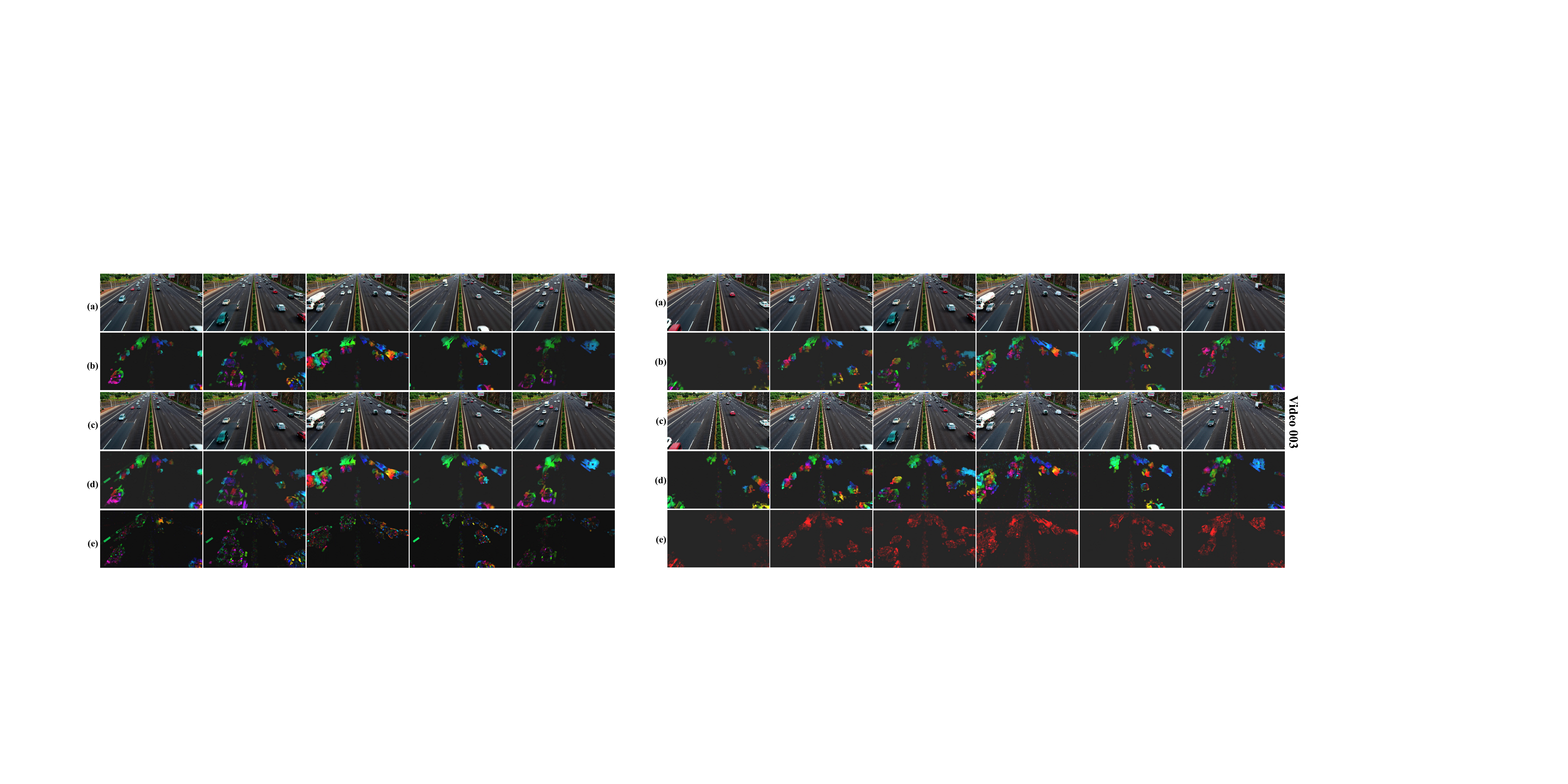}
\caption{Visualizations sampled from Combined-VSR. The rows from (a) to (e) denote the reference video, optical flow of the reference video, the SR video, optical flow of the SR video, and the temporal inconsistency video. Columns from left to right display consecutive frames sampled from the videos, with the sampling frame gap equal to the frame rate (FPS). The MOS/Quality Prediction of VSFA/MBVQA/\textcolor{red}{TIG-SVQA}(Ours) are as follows: 0.15073/0.89866/0.45200/\textcolor{red}{0.14178}.} 
\label{fig:5}
\end{figure*}
\subsection{Visualization}
As illustrated in Figure~\ref{fig:5}, we visualize the optical flows of super-resolution (SR) videos alongside their corresponding reference videos and derive temporal inconsistency information by calculating the differences between them. In these visualizations, color encodes motion direction, while brightness reflects motion magnitude.

Unlike traditional motion measurement methods such as optical flow—which capture standard motion cues but are often masked by complex scene content—our temporal inconsistency visualizations provide a more direct and interpretable representation of distortions caused by temporal inconsistencies in SR videos.

Furthermore, when comparing our predicted quality scores against those generated by two state-of-the-art video quality assessment methods, we observed that our predictions were more strongly correlated with the ground-truth Mean Opinion Scores (MOS), demonstrating the effectiveness of our approach in capturing perceptual quality.
\subsection{Ablation on Loss Function}
In our experiments, we attempted to use both L1 loss and MSE loss, and added SRCC as a penalty term based on these two losses. The calculation is as follows: 
\begin{equation}
\begin{aligned}
    &\ \mathcal{L}_{L_{1}\&SRCC} = \mathcal{L}_{L_{1}} + (1-SRCC) \\
    &\ \mathcal{L}_{MSE\&SRCC} = \mathcal{L}_{MSE} + (1-SRCC).
\end{aligned}
\end{equation}
The experimental results are shown in Table \ref{tab:table3}, where each loss was evaluated by three random splits of the dataset (these three repeated tests differ from the ten repetitions mentioned in the main text, leading to slight differences in the average results). It can be observed that the MSE \& SRCC loss achieved the best performance, while the difference between the effects of L1 loss and MSE loss was minimal. 
\begin{table}[t]
  \centering
  \caption{Validity of different loss functions.}
  \resizebox{1.0\linewidth}{!}{
    \begin{tabular}{c|cccc}
    \hline
    Loss  & SRCC$\uparrow$ & PLCC$\uparrow$ & KRCC$\uparrow$ & RMSE$\downarrow$ \\
    \hline
    L1 Loss & 0.926  & 0.929  & 0.768  & 0.100  \\
    L1\&SRCC Loss & 0.941  & 0.942  & 0.794  & 0.087  \\
    MSE Loss & 0.929  & 0.933  & 0.771  & 0.092  \\
    MSE\&SRCC Loss & \textbf{0.944}  & \textbf{0.945}  & \textbf{0.798}  & \textbf{0.082}  \\
    \hline
    \end{tabular}}
  \label{tab:table3}%
\end{table}%

We further performed a T-test on the SRCC results of different loss functions, with the outcomes shown in the Table \ref{tab:table4}. The P-values between $L_{1}$ and $L_{1} \& SRCC$ is 0.01919, $L_{1}$ and $MSE$ is 0.12025, $L_{1}$ and $MSE \& SRCC$ is 0.00368, $L_{1}$ and $SRCC \& MSE$ is 0.03986, $L_{1} \& SRCC$ and $MSE \& SRCC$ is 0.58564, $MSE$ and $MSE \& SRCC$ is 0.00687. 
\begin{table}[t]
  \centering
  \caption{Results of the two sample T-test performance between SRCC values obtained by different loss functions. }
  \resizebox{1.0\linewidth}{!}{
    \begin{tabular}{c|cccc}
    \hline
    Loss  & \makecell{L1\\Loss} & \makecell{L1-SRCC\\Loss} & \makecell{MSE\\Loss} & \makecell{MSE\&SRCC\\Loss} \\
    \hline
    L1 Loss & 0     & -1    & 0     & -1 \\
    L1-SRCC Loss & 1     & 0     & 1     & 0 \\
    MSE Loss & 0     & -1    & 0     & -1 \\
    MSE\&SRCC Loss & 1     & 0     & 1     & 0 \\
    \hline
    \end{tabular}}
  \label{tab:table4}%
\end{table}%
\section{Generalization on Real-world Scenarios}
For real-world distortions, we evaluated the generalization capability of our method on two UGC datasets: KoNViD-1K~\cite{konvid-1k} and LIVE-VQC~\cite{LIVEVQC}. As shown in Figure~\ref{fig:6}, for these UGC scenarios, temporal inconsistency information is estimated directly from motion cues, specifically by utilizing optical flow. This is because, in real-world scenarios, reference videos are typically unavailable, making it infeasible to compute inconsistency based on motion differences. Instead, motion information serves as a practical proxy to approximate temporal inconsistency. The effectiveness of this approximation is empirically supported by the correlation shown in Figure \ref{fig:6}, demonstrating that optical flow can provide meaningful perceptual cues for UGC videos.

KoNViD-1K consists of 1,200 videos with 8-second durations, 540p resolution, and frame rates between 24 and 25 fps, focused on natural distortions. LIVE-VQC includes 585 videos, each 10 seconds long, with varying resolutions (240p to 1080p) and a frame rate of 30 fps, also targeting natural distortions.
\subsection{Generalization on UGC Datasets}
For each UGC dataset, we randomly split the dataset into $60\%$ training, $20\%$ validation, and $20\%$ testing, repeating the process 10 times and reporting median results.
\begin{table*}[t]
    \centering
    \caption{Performance comparison of our methods against competing UGC VQA methods on two widely applied UGC datasets (i.e., LIVE-VQC and KoNViD-1K). All the models were trained individually on each dataset. The best performance is highlighted.}
     \resizebox{1.0\linewidth}{!}{
    \begin{tabular}{c|c|ccccccccccc}
    \hline
        Datasets & Metrics & VSFA & TLVQM & VIDEVAL & VISION & SimpleVQA & VIQE & 2Bi-VQA & STFR & MBVQA & FAST-VQA & \textbf{TIG-SVQA}  \\ \hline
        \multirow{2}{*}{KoNViD-1K} & PLCC$\uparrow$ & 0.755  & 0.764  & 0.772  & 0.632  & 0.798  & 0.638  & 0.835  & 0.826  & 0.821  & 0.827  & \textbf{0.844}   \\ 
        ~ & SRCC$\uparrow$ & 0.788  & 0.760  & 0.774  & 0.598  & 0.792  & 0.628  & 0.815  & 0.822  & 0.829  & 0.827  & \textbf{0.842}   \\  \hline
        \multirow{2}{*}{LIVE-VQC} & PLCC$\uparrow$ & 0.663  & 0.432  & 0.752  & 0.689  & 0.775  & 0.694  & 0.832  & 0.805  & 0.831  & 0.737  & \textbf{0.838}   \\ 
        ~ & SRCC$\uparrow$ & 0.640  & 0.450  & 0.751  & 0.676  & 0.740  & 0.660  & 0.761  & 0.801  & 0.749  & 0.687 & \textbf{0.808}   \\ \hline
    \end{tabular}}
    \label{t3} 
\end{table*}
\begin{table}[t]
\centering
\caption{Cross-dataset testing of our method against two competing methods. The best performance is highlighted.}
\resizebox{1.0\linewidth}{!}{
\begin{tabular}{c|c|cc|cc|cc} \hline
\multicolumn{2}{c|}{Methods} & \multicolumn{2}{c|}{\textbf{TIG-SVQA}} & 
\multicolumn{2}{c|}{2Bi-VQA} & \multicolumn{2}{c}{MBVQA} \\ \hline 
Train & Test & SRCC$\uparrow $& PLCC$\uparrow$ & SRCC$\uparrow$ & PLCC$\uparrow$ & SRCC$\uparrow$ & PLCC$\uparrow$ \\ \hline
Combined-VSR & KoNViD-1K & \textbf{0.701} & 0.659 & 0.678 & \textbf{0.665} & 0.611 & 0.601\\
Combined-VSR & LIVE-VQC & \textbf{0.630} & \textbf{0.687} & 0.615 & 0.672 & 0.627 & 0.638 \\
KoNViD-1K & LIVE-VQC & \textbf{0.804} & 0.813 & 0.768 & \textbf{0.840} & 0.636 & 0.649 \\
KoNViD-1K & Combined-VSR & \textbf{0.565} & \textbf{0.573} & 0.404 & 0.391 & 0.526 & 0.544 \\
LIVE-VQC & KoNViD-1K & 0.733 & 0.752 & \textbf{0.776} & \textbf{0.759} & 0.674 & 0.660 \\
LIVE-VQC & Combined-VSR & \textbf{0.406} & \textbf{0.379} & 0.320 & 0.322 & 0.345 & 0.371 \\ \hline
\end{tabular}}
\label{t4}
\end{table}

Table \ref{t3} shows that TIG-SVQA consistently outperforms other VQA methods on both the KoNViD-1K and LIVE-VQC datasets. On KoNViD-1K, TIG-SVQA achieves the highest PLCC (0.844) and SRCC (0.842), matches the models like MBVQA, 2Bi-VQA, and STFR. Similarly, on LIVE-VQC, TIG-SVQA leads with a PLCC of 0.838 and SRCC of 0.808, indicating its robust capability in assessing UGC video quality.
\subsection{Validity on Cross datasets}
As shown in Table~\ref{t4}, the cross-dataset evaluation is specifically designed to validate the proposed model’s generalization ability across different distortion types—that is, training on super-resolved (SR) videos with distortions derived from enhancement processes and testing on user-generated content (UGC) videos with real-world degradations, and vice versa. Under this rigorous setting, TIG-SVQA consistently outperforms two strong baselines, 2Bi-VQA and MBVQA. When trained on the Combined-VSR dataset and tested on the KoNViD-1K and LIVE-VQC datasets, TIG-SVQA achieves better performance, indicating strong generalization from SR to real-world distortions. Conversely, when trained on UGC datasets and tested on Combined-VSR, TIG-SVQA still demonstrates competitive results. These findings highlight the superior cross-distortion generalizability of our model. 
\subsection{Validity on Fused datasets}
To further validate the applicability of TIG-SVQA for both natural and super-resolution distortions, we merged the Combined-VSR and UGC datasets for model training. Table \ref{t5} presents a comparison of the TIG-SVQA with the other state-of-the-art VQA methods on the fused dataset. Both SRCC and PLCC values indicate that TIG-SVQA achieves the highest performance across all aspects, demonstrating its robustness in video quality assessment. In contrast, while MBVQA performs well, its scores fall short of TIG-SVQA's precision. The 2Bi-VQA exhibits significantly lower performance, highlighting its limitations in effectively assessing video quality across different distortion scenarios compared to TIG-SVQA and MBVQA.
\begin{table}[t]
\centering
\caption{Fused-dataset evaluation of our method against two competing methods. The best performance is highlighted.}
\resizebox{1.0\linewidth}{!}{
\begin{tabular}{c|cccc|cccc}
\hline
Datasets & \multicolumn{4}{|c|}{Combined-VSR + KoNViD-1K} & \multicolumn{4}{c}{Combined-VSR + LIVE-VQC} \\ \hline
Methods & SRCC$\uparrow$ & PLCC$\uparrow$ & KROCC$\uparrow$ & RMSE$\downarrow$ & SRCC$\uparrow$ & PLCC$\uparrow$ & KROCC$\uparrow$ & RMSE$\downarrow$ \\ \hline
2Bi-VQA & 0.534 & 0.445 & 0.381 & 1.939 & 0.511 & 0.413 &  0.358 & 2.022 \\
MBVQA & 0.898 & 0.927 & 0.733 & 0.087 & 0.871 & 0.885 & 0.695 & 0.098 \\
\textbf{TIG-SVQA} & \textbf{0.928} &\textbf{0.938} & \textbf{0.775} & \textbf{0.086} & \textbf{0.949} & \textbf{0.944} & \textbf{0.809} & \textbf{0.080} \\ \hline
\end{tabular}}
\label{t5}
\end{table}

\subsection{Ablation of Inconsistency Guidance on UGC Datasets}
\label{s1}
Table \ref{tab:table8} highlights the significant improvements achieved by incorporating inconsistency guidance in video quality assessment models when applied to UGC datasets, KoNViD-1K and LIVE-VQC. Models without inconsistency guidance exhibit relatively weaker performance, particularly when relying solely on coarse-grained features, which limits their ability to accurately predict video quality. In contrast, integrating inconsistency guidance leads to substantial gains in SRCC and PLCC scores across datasets, especially when combining fine-grained and coarse-grained features. Furthermore, it is worth noting that models using only fine-grained features generally outperform those using only coarse-grained features. This may be attributed to the nuanced nature of temporal inconsistency, where local features play a more critical role in capturing subtle variations.
\begin{table}[t]
\centering
\caption{The ablation study on two UGC datasets: KoNViD-1K and LIVE-VQC. ``w/o" indicates ``without". ``Fine" and ``Coarse" represent the use of spatial features extracted solely by the Fine Scene Spatial Extractor and the Coarse Scene Spatial Extractor, respectively.}
\resizebox{1.0\linewidth}{!}{
\begin{tabular}{c|cc|cccc}
\hline
 & \multicolumn{2}{c|}{Dataset} & \multicolumn{2}{c}{KoNViD-1K} & \multicolumn{2}{c}{LIVE-VQC} \\
\hline
Model & Fine & Coarse & SRCC$\uparrow$ & PLCC$\uparrow$ & SRCC$\uparrow$ & PLCC$\uparrow$ \\
\hline
\multirow{3}{*}{\makecell{w/o \\ Inconsistency \\ Guidance}} 
 & $\checkmark$ & $\times$ & 0.772 & 0.787 & 0.769 & 0.818 \\
 & $\times$ & $\checkmark$ & 0.623 & 0.664 & 0.658 & 0.771 \\
 & $\checkmark$ & $\checkmark$ & 0.804 & 0.816 & 0.770 & 0.832 \\
 \hline
\multirow{3}{*}{\makecell{with \\ Inconsistency \\ Guidance}} 
 & $\checkmark$ & $\times$ & 0.780 & 0.818 & 0.807 & 0.810 \\
 & $\times$ & $\checkmark$ & 0.709 & 0.720 & 0.671 & 0.808 \\
 & $\checkmark$ & $\checkmark$ & \textbf{0.842} & \textbf{0.844} & \textbf{0.808} & \textbf{0.838} \\
\hline
\end{tabular}}
\label{tab:table8}
\end{table}

\subsection{Limitations}
Considering that optical flow computation in the proposed method is time-consuming, we measured the average inference time of FAST-VQA (0.19s), MBVQA (9.12s), and TIG-SVQA (55.26s). For TIG-SVQA, the optical flow computation alone takes approximately 54.45 seconds per video, whereas the actual model inference requires only 0.81 seconds. This clearly indicates that optical flow computation is a bottleneck in the current version.

Currently, the optical flow estimation in TIG-SVQA is implemented using the Farneback method~\cite{farneback2003two}, a classical non-deep-learning algorithm that is relatively lightweight and easy to deploy. Although lightweight deep-learning-based models like LiteFlowNet~\cite{hui2018liteflownet} have been proposed, our empirical tests show that they do not significantly improve inference efficiency due to the additional overhead of model loading and GPU computation. As a result, the Farneback method remains a more practical choice in the current implementation. In the future, we plan to investigate alternative temporal inconsistency descriptors that reduce reliance on dense optical flow computation.
\section{Conclusion}
This paper presents TIG-SVQA, a novel framework for super-resolution video quality assessment. We begin by quantifying temporal inconsistency and demonstrating its strong correlation with human perception. Using this insight, TIG-SVQA improves the accuracy of perceptual prediction under the guidance of temporal inconsistency information. The proposed method first introduces an Inconsistency Highlighted Spatial Module that extracts coarse- and fine-grained inconsistency features, incorporating a newly designed DW-SA-T block. Then, we propose an Inconsistency Guided Temporal Module, which consists of consistency-aware fusion and informative filtering stages. A visual memory capacity block adaptively allocates temporal segments in the first stage, while the filtering process further emphasizes quality-oriented features in the second stage. Extensive experiments on both single-frame and multi-frame SR video quality datasets demonstrate the superior performance of TIG-SVQA. 
\section{Acknowledgments}
This work was in part supported by the National Natural Science Foundation of China under Grant 62371017.
\bibliography{aaai2026}


\end{document}